\documentclass[10pt, a4paper, onecolumn]{article}

\usepackage{arxiv}

\usepackage[utf8]{inputenc} 
\usepackage[T1]{fontenc}    
\usepackage{hyperref}       
\usepackage{url}            
\usepackage{booktabs}       
\usepackage{amsfonts}       
\usepackage{nicefrac}       
\usepackage{microtype}      
\usepackage{lipsum}
\usepackage{amsmath}
\usepackage{cite}
\usepackage{amssymb}
\usepackage{verbatim}
\usepackage{enumerate}
\usepackage{mdwlist} 
\usepackage{color}

\usepackage{algpseudocode}
\usepackage{algorithmicx}
\usepackage{lineno}
\usepackage{framed,multirow}
\usepackage[final]{pdfpages}
\usepackage{latexsym}
\usepackage{wrapfig}

\usepackage{url}
\usepackage{xcolor}
\usepackage{graphicx}
\usepackage{subfigure}
\usepackage{verbatim}
\usepackage{bm}
\usepackage{authblk}
\usepackage{tikz}
\usepackage[space]{grffile}
\usetikzlibrary{shapes,arrows}

\title{Structure and Performance of Fully Connected Neural Networks: Emerging Complex Network Properties}

\author[1]{Leonardo F. S. Scabini}
\author[1]{Odemir M. Bruno}
\affil[1]{\small{S\~{a}o Carlos Institute of Physics, University of S\~{a}o Paulo \protect\\ Trabalhador São-carlense avenue, 400, postal code 13560-970, S\~{a}o Carlos, São Paulo, Brazil \protect\\ $\{$scabini,bruno$\}$@ifsc.usp.br}}

\begin{document}
\maketitle

\begin{abstract}
Understanding the behavior of Artificial Neural Networks is one of the main topics in the field recently, as black-box approaches have become usual since the widespread of deep learning. Such high-dimensional models may manifest instabilities and weird properties that resemble complex systems. Therefore, we propose Complex Network (CN) techniques to analyze the structure and performance of fully connected neural networks. For that, we build a dataset with 4 thousand models and their respective CN properties. They are employed in a supervised classification setup considering four vision benchmarks. Each neural network is approached as a weighted and undirected graph of neurons and synapses, and centrality measures are computed after training. Results show that these measures are highly related to the network classification performance. We also propose the concept of Bag-Of-Neurons (BoN), a CN-based approach for finding topological signatures linking similar neurons. Results suggest that six neuronal types emerge in such networks, independently of the target domain, and are distributed differently according to classification accuracy. We also tackle specific CN properties related to performance, such as higher subgraph centrality on lower-performing models. Our findings suggest that CN properties play a critical role in the performance of fully connected neural networks, with topological patterns emerging independently on a wide range of models.
\keywords{artificial neural networks \and complex networks \and computer vision \and machine learning} 
\end{abstract}


\section{Introduction}

Artificial Neural Networks (ANN) are in a non-stopping ascension since the introduction of deep learning, and many successful applications have been developed \cite{najafabadi2015deeplapplications}. The state-of-the-art on various applications is constantly pushed forward by innovative neural architectures, training mechanisms, and big data collections. The fast advances in these techniques and hardware technology make it possible to create deeper and more complex models. However, this phenomenon leads to black-box approaches becoming usual in many applications, as the study of ANN fundamentals does not follow the same rhythm. For instance, the lack of understanding of its internal details \cite{basu2016theoretical} causes some intriguing inexplicable properties to be observed in deep convolutional networks \cite{nguyen2015deep}, one of the most popular ANN architectures for Computer Vision. Experiments have shown that imperceptible input perturbations can drastically impact performance, a phenomenon also known as adversarial examples \cite{goodfellow2014explaining}. It is also possible to create entirely unrecognizable images, but that convolutional networks strongly believe are recognizable objects \cite{nguyen2015deep}. These performance instabilities limits research not only on neural network functioning, but also on network optimization such as better synapse initialization, construction, architecture search, and improved fitting mechanisms. Improving our knowledge of their properties will then lead to Explainable Artificial Intelligence (XAI) \cite{adadi2018peeking} and also more robust and accurate models. Therefore, understanding the functioning of ANNs is recently one of the main topics within the field, which has become a true interdisciplinary research uniting computer scientists with neuroscientists, psychologists, mathematicians and physicists.


Some of the weird behaviors observed on ANNs are compatible with complex systems, which are usually non-linear, hard to predict, and sometimes sensitive to small perturbations. One interesting approach then to study such systems is through Complex Networks (CN), or Network Science, a research field that emerges from graph theory, physics, and statistics. Its main goal is to study large networks derived from natural, complex processes, which correlate to various neural networks such as biological ones. For instance, it was shown the existence of the small-world CN phenomenon in the neural network of the \textit{Caenorhabditis elegans} worm \cite{scalefreeCN}. Various works then found such topological patterns also in human neural networks, like from magnetoencephalography recordings \cite{stam2004functional}, analysis of cortical connections \cite{sporns2004small} and of the brain stem medial reticular formation  \cite{humphries2005brainstem}. In this context, CN principles have helped to understand the origin of dynamic brain activity in anatomical circuits at various levels of scale \cite{sporns2011human}.

In the case of ANN, few works have approached them from CN principles. In  \cite{stauffer2003efficient}, Hopfield networks are simulated as scale-free CNs, reducing the number of neuron connections and reducing the size and computational cost while keeping tolerable performance. A work from 2004 \cite{torres2004influence} shows that attractor networks with scale-free topology have a greater capacity to store and retrieve binary patterns than highly random Hopfield networks with the same number of synapses. The relative performance of the scale-free system increases with increasing values of its power-law exponent. More recently (2016) \cite{erkaymaz2016impact} and after the deep learning diffusion, authors proposed a small-world ANN that has better performance than traditional structure in the task of a diabetes diagnosis. In a 2020 work \cite{testolin2020deep}, CNs are employed for the analysis of 5-layer deep-belief networks, and results suggest correlations between topological properties and functioning. Another 2020 work \cite{zambra2020emergence} explores the emergence of network motifs, i.e., a group of neurons with a distinct connection pattern, on a small Multilayer Perceptron (MLP) ANN. They considered different number of neurons on this architecture applied on small synthetic data, and concludes that specific motifs emerge during training. In a 2021 work \cite{florindo21}, CNs were used to analyze the output of convolutional neural networks in texture images. This work also corroborates with the existence of CN information within ANNs, which they consider for enhancing the model's classification power for texture analysis.

Although these works suggest CN properties within ANNs, much still needs to be done to uncover its properties. Moreover, these works do not consider a reliable number of neural networks to account for the wide possible states of random synapse organizations. It is widely known that different weight initialization can cause drastic changes in neural network behavior, such as some specific distributions which may make training particularly effective \cite{frankle2018lottery}.

The present work approaches fully connected multilayer neural networks as CNs, focusing on neuronal topological properties. We considered deep MLP-like networks applied for supervised classification on vision tasks, which is also one of the most diffused ANN areas. Our analysis consider a large neural network population (4 thousand models), with different initial synapses, trained on four vision benchmarks, which we make available as a new dataset for studying ANNs \footnote{\label{github}\url{https://github.com/scabini/MLP-CN_dataset}}. They are characterized through CN centrality measures computed from its hidden neurons, which are also available in the repository. The following sections of the paper are divided by: Section \ref{sec:theory} gives the background fundamentals on ANNs and CNs; Section \ref{sec:method} describe the proposed methodology for building the neural network dataset and analyzing it through CNs; Section \ref{sec:results} shows and discuss our analyzes and obtained results; and finally, Section \ref{sec:conclusion} concludes the work with the overall findings, the most critical remarks, and future works.

\section{Theoretical Background}\label{sec:theory}

This section introduces the fundamentals of ANNs and CNs, focusing on models and techniques employed in this work.

\subsection{Artificial Neural Networks}


After the first neuroscience discoveries about the brain, scientists started to study artificial models. Around 1943, from the work ``\textit{A logical calculus of the ideas immanent in nervous tissue}" \cite{mcculloch1943logical}, the first ideas of artificial neurons are established.
In 1958, Rosenblatt introduced the Perceptron \cite{rosenblatt1958perceptron}, the first artificial neuron capable of performing pattern recognition tasks after training. It can be defined as a function $\phi(\vec{X})=y$, which computes $b + \sum x_i w_i$, where $w_i$ are learnable weights and $x_i$ are input elements of $\vec{X}$. The neuron's output results from its activation function $\phi$, which is the Heaviside step function (threshold function) that returns 1 if $y > 1$, and 0 otherwise. The training process consists of adjusting each weight $w_i$ in a supervised fashion, i.e., the desired output is known for the training set. Given a training sample, the new weight is adjusted by 
\begin{equation}
w_i^{t+1} = w_i^t + \alpha (\bar{y} - y)x_i    
\end{equation}
where $\bar{y}$ is the expected output for the input vector $\vec{X}$ and $\alpha$ is the learning rate, that controls the update magnitude. By iterative repeating this process for $e$ epochs in a given training set, the Perceptron tends to find a solution for a binary problem if it is linearly separable.


ANNs, in turn, are composed of a chain of connected neurons usually organized in $d$ layers. Theoretically, a network extensive enough could fit any function, which is one reason for attracting the scientific community's attention.
One of the most traditional ANN models is the MLP \cite{bishop1995neural}, usually composed of $d=3$ layers: input, hidden layer, and output. These layers are fully connected with the next and previous layers, and no connections exist between neurons of the same layer. A layer can be represented by a matrix $W_{ij}^{a+1}$ that describes the weights of connections between the $i$ neurons of layer $a$ with the $j$ neurons of layer $a+1$. The output of a layer is then computed by matrix operations
\begin{equation}
   \phi(\vec{X} \cdot W)=\vec{Y} 
\end{equation}
where $\vec{Y}$ is a vector with the output of each layer's neurons. The activation function $\phi$ is usually a non-linear operator. The process is then repeated to each of the $d$ layers, using the output of the previous layer ($\vec{Y^{a-1}}$) as input to the next one, multiplying to the corresponding weights $\phi(\vec{Y^{a-1}} \cdot W^a )=\vec{Y^a}$, where $j$ and $l$ are neuron indices from layer $a-1$ and $a$. At the last layer, the final output $\vec{Y^{d}}$ is obtained by a different activation function $\phi$, corresponding to the problem one needs to address (e.g., linear for binary classification, or softmax for multiclass tasks). The whole model can then be approached as a collection of layers $\aleph=\{L_0,...,L_d\}$, each one with a set of associated weights $W$, and an activation function $\phi$. Between the decades of 1970 and 1980 \cite{rumelhart1988learning} advances on these networks' training through backpropagation helped consolidate the multilayer approach.
In most cases of supervised classification this training technique is prevalent until today and consists of propagating the last layer's error ($ \vec{Y^{d}} - \vec{X}$) backward, updating weight matrices $W^a$ layer-by-layer similarly to the Perceptron approach.

Although the theory behind fully connected networks supports an ideal pattern recognizer hypothesis, its performance remained similar to other statistical and logical methods between 1990 and 2000. This stagnation happened due to the lack of efficient training techniques for deeper and larger networks. This training approach, widely known as deep learning, faced difficulties such as the natural occurrence of overfitting due to the vast number of parameters (synapses) on larger models. The most common model of this kind is the deep feedforward network, which propagate information in a single direction, and is similar to the MLP but with multiple hidden layers ($d>3$). A similar model (deep belief network) gained attention in 2006 \cite{hinton2006fast} when authors showed that the unsupervised training of layers as a way of weight initialization facilitates the use of deeper models. It is essential to notice that deeper networks also gained more attention due to the computer power of graphic processing units (GPU) \cite{raina2009large}. In linear algebra, especially in vector and matrix operations, GPU is superior to CPU processing, allowing faster training of more significant amounts of neurons and layers. In 2012 the deep convolutional network AlexNet \cite{krizhevsky2012imagenet} was introduced, with clever training techniques and GPU implementation, demonstrating high discriminatory power on the ImageNet object recognition challenge \cite{imagenet}. Since then, deep models have become the predominant approach to computer vision, and several deep architectures have been proposed. However, almost all modern ANNs still contain some fully connected parts resembling the early MLP. Moreover, it has been shown recently that pure MLP models (with additional tricks) can be employed in deep learning tasks  with state-of-the-art performance \cite{tolstikhin2021mlp,liu2021pay}.

\subsubsection{Parameters and training}

This work focuses on fully connected, feedforward ANNs that resemble the MLP but with multiple hidden layers. Although this architecture is simple, there is a wide range of construction blocks, such as activation functions \cite{glorot2011relu}, synapse initializers  \cite{glorot2010understanding}, etc. For instance, for synapse initialization, which is the process of creating a random initial distribution to compose the weight matrices $W^a$, there are optimized techniques such as Glorot-normal \cite{glorot2010understanding} rather than random uniform distributions. For neuron activation functions, most methods usually intend to project the inputs non-linearly. We can mention the Rectifier-Linear-Unit $f(x) = x^+ = max(0,x)$ \cite{glorot2011relu}, an activation function that became popular after the widespread of deep models due to its simplicity and efficiency. However, these choices usually depend on the problem addressed or the architecture style, and they are defined mostly by empirical tests. 

Moreover, beyond the ANNs construction stage, there are many training particularities, such as data preparation/normalization, optimizers, learning parameters, and stopping criteria. Current vision models usually normalize data according to the ImageNet standards \cite{imagenet}, which consists of a z-score normalization that centers the data on 0 (subtracts the average) and makes the variance equals to 1. For training, popular optimizers are backpropagation \cite{rumelhart1988learning} approaches such as the Stochastic Gradient Descent (SGD) with momentum \cite{sutskever2013importance}. 
Each optimizer has his hyperparameters that need to be tuned. The number of training epochs $e$ is probably the most empirical one, where the user needs to supervise training performance to decide when to stop it.

In short, the ANNs development requires extensive architecture engineering and empirical tests that also is hardware demanding. This search is not random, i.e., there are good practices in the field and known configurations that work in a common scenario. However, they do not ensure the model will always work as expected. Moreover, other properties may also cause instability and vulnerability on neural networks, such as adversarial samples \cite{goodfellow2014explaining}. These drawbacks are another motivation for better understanding the internal dynamics of ANNs.

\subsection{Complex Networks}

The CN research field, or Network Science, emerged from the interaction between different areas such as graph theory, physics, and statistics. Its main goal is to study large networks derived from natural, complex processes. A classic example is the internet, a global network composed of many connected devices transmitting information worldwide. Furthermore, nature and modern society are filled with similar network-like systems, such as cell compound interaction, traffic flow, social networks, etc. 

It is not simple to analyze complex systems due to the difficulty of deriving their collective behavior from a knowledge of the system's components \cite{barabasi2016network}. Initially, researchers believed that most networks from natural complex systems had random topology \cite{randomCNevolution}. However, researchers eventually found structural patterns, leading to better models to describe them. The most diffused ones are the scale-free \cite{scalefreeCN}, and small-world \cite{smallworldCN} networks. Therefore, these breakthroughs opened a new door for pattern recognition, where CNs are employed as a tool for the modeling and characterization of natural phenomena. These concepts are now applied in a broad range of areas \cite{aplicacoesRC}.

A network can be defined by a tuple $R=(V, E)$, where $V=\{v_1, ..., v_n\}$ represents its vertices and $E=\{e(v_i,v_j)\}$ the connections between vertex pairs. Each connection may be represented in several forms, and the most common approach is undirected, unweighted edges. That is,  $e(v_i,v_j)=1$ if $v_i$ and $v_j$ are linked (0 otherwise), and $e(v_i,v_j) = e(v_j,v_i)$. 
Weighted edges has a real value (not binary) associated with the connection between the vertices, i.e. $e(v_i,v_j) \in \mathbb{R}$, which may represent some kind of distance metric, connection strength or similarity. A network can also be represented by an adjacency matrix $A=[a(v_i,v_j)]$ where $i,j=1,...,n$ and $a(v_i,v_j)=1$ represents the existence of a link between $v_i$ and $v_j$, and 0 represents no connection. 

\subsubsection{CN models and measures}\label{sec:CNmeasures}

From CN vertices and edges, it is possible to calculate a wide range of metrics to quantify its topology \cite{cnusp}, each of which evidences a different aspect from the network. The degree is the most diffused measure, and can be calculated for each vertex $v_i$ by simply summing its adjacency matrix row ($\sum_j a(v_i, v_j)$), representing the number of edges leaving $v_i$. In weighted networks, this sum considers real values corresponding to each edge weight, and it is usually called weighted degree or strength
\begin{equation}\label{eq:s}
    s(v_i) = \sum\limits_{\forall v_j \in V} e(v_i, v_j).
\end{equation}

The degree is the basis to describe several network properties, such as the scale-free phenomenon \cite{scalefreeCN}. In these networks, the degree distribution follows a power law distribution $p(s) \approx s^{- \gamma}$, where $\gamma$ varies according to the network structure. This topological pattern highlights hubs' existence, i.e., some vertices have a very high degree and play a critical role in the network's internal functioning. In contrast, most of the other vertices have a low degree. Several works have found the scale-free property in real-world networks from the most diverse domains, usually with $\gamma \approx 3$.

It is also possible to calculate second order measures considering the properties of the node neighbors. This approach is based on the assortativity principle, i.e., that nodes tend to connect to other similar nodes. For instance, the average neighbor strength (for weighted networks) \cite{barrat2004architecture} of a node $v_i$ simply considers the mean strength of other nodes connected to it
\begin{equation}\label{eq:sw}
    s^w_{nn}(v_i) = \frac{1}{s(v_i)} \sum\limits_{v_j \in N(v_i)} e(v_i, v_j) s(v_j) 
\end{equation}
where $N(v_i)$ define the set of nodes connected to $v_i$. Another more complex measure following this principle is the second order centrality \cite{kermarrec2011second}, which computes the standard deviation of the return times to a given node $v_i$ during a perpetual random walk on the network. In practice, it can be solved analytically, i.e., without the simulation of a random walk 
\begin{equation}\label{eq:so}
     so(v_i) = \sqrt{2 \sum\limits_{v_j \in V } M(v_j,v_i) - n(n+1) }
\end{equation}
where a classical discrete time Markov chain $M$ is employed (further details can be consulted on \cite{kermarrec2011second}). Another walk-based measure is the subgraph centrality \cite{estrada2005subgraph}, which considers all weighted closed walks, of all sizes, starting and ending at a given vertex. Each closed walk is associated with a connected subgraph, and the weights decrease with path length. This measure can be computed spectrally using the eigenvalues and eigenvectors of the adjacency matrix
\begin{equation}\label{eq:sg}
    sg(v_i) = \sum\limits^n_j (u_j^{v_i})^2 e^{\lambda_j}
\end{equation}
where $u_j$ is an eigenvector corresponding to the eigenvalue $\lambda_j$ of $A$.

In a similar fashion, it is possible to find graph cliques \cite{bron1973algorithm,cazals2008note}, i.e., a subset of nodes that are all adjacent. These objects can also be interpreted as a complete subgraph of the network. A maximum clique is the most extensive clique under this condition. It is possible to compute the number of maximum cliques that a vertex participates

\begin{equation}\label{eq:mc}
\begin{split}
Q(G) = [M_1,...,M_J] \\
    mc(v_i) = \sum_{j}^{} \left\{\begin{array}{rcl}
    1,& \textnormal{if } v_i \in M_j \\    
    0,& \textnormal{otherwise}.
    \end{array}\right. 
\end{split}
\end{equation}
where $Q$ is the set of maximum cliques in $G$, and $M_j$ are sets of nodes representing each maximum clique. Finding maximum cliques in a graph is a NP-complete problem. The search is usually performed by depth-first search algorithms and pruning the search tree, resorting to well chosen pivoting nodes \cite{cazals2008note}.

Small-world networks \cite{smallworldCN} are models whose principal property is the facility for information spreading. In other words, these networks have a short average shortest-path distance where vertices can be reached easily (through few steps). Another essential property of these networks is their high clustering coefficient, which describes the level of interconnectivity between neighboring vertices. This measure counts the number of triangles (fully connected triple) that occurs between a vertex and its neighbors by $c(v_i) = \frac{2 \bigtriangledown (v_i)}{k(v_i)(k(v_i)-1)}$, where $\bigtriangledown (v_i)$ is the number of triangles in which $v_i$ participates, or simply the number of edges between the neighbors of $v_i$. 
It can also be computed for bipartite networks, also referred as bipartite local clustering \cite{latapy2008basic}
\begin{equation}\label{eq:bc}
  bc(v_i) = \frac{\sum\limits_{v_j \in N(N(v_i))} pc(v_i, v_j)}{|N(N(v_i))|} 
\end{equation}
where $N(N(v_i))$ are the second order neighbors of $v_i$ (neighbors of the neighbors, excluding $v_i$), and $pc(v_i, v_j) = \frac{|N(v_i) \cap N(v_j)|}{max(|N(v_i)| , |N(v_j)|}$ is the pairwise cluster coefficient between nodes $v_i$ and $v_j$.

Path-based measures are also important for characterizing information flow, such as how fast a node communicates to the others. The harmonic centrality \cite{boldi2014axioms} considers this principle and computes the sum of the reciprocal of the shortest path distances from all the vertices to a given vertex
\begin{equation}\label{eq:hc}
    hc(v_i) = \sum\limits_{j \in V } \frac{1}{d(v_j,v_i)}
\end{equation}
where $d(v_j,v_i)$ is the Dijkstra’s algorithm \cite{dijkstra1959note} for computing the shortest distance between two nodes $v_i$ and $v_j$, considering the edge values for weighting the calculation. However, this computation ignores all other than the shortest paths. In some scenarios, information may flow throughout all the network paths. Therefore, some measures relaxes the shortest-path definition by including all paths and weighting according to its length. For instance, the current flow (CF) approach \cite{brandes2005centrality}, inspired by the electric flow on circuits. The idea is to consider the network as a resistor network where edges are resistors and vertices junctions between resistors. The electric network analogy is achieved by computing the inverse Laplacian matrix (see \cite{brandes2005centrality} for further details). In this scenario, a possible measure is the CF closeness
\begin{equation}\label{eq:cfc}
    CF_c(v_i) = \frac{n - 1}{\sum\limits_{v_j  \in V} p_{v_i, v_j}(v_i) - p_{v_i, v_j}(v_j)}
\end{equation}
where $p_{v_i, v_j}(v_i) - p_{v_i, v_j}(v_j)$ corresponds to the effective resistance, i.e., the alternative distance metric between the two nodes which consider all possible paths on the eletric network.

Beyond the CN properties and measures we describe here, the literature is rich and heterogeneous on other approaches \cite{cnusp}. We limited our scope here due to several factors we discuss further in the following section.

\section{Methodology}\label{sec:method}

In this work, we analyze the internal CN properties of fully connected neural networks and their correlation to classification performance on vision tasks. This architecture is considered one of the most diffused models since early neural networks studies, and it is still popular among modern deep methods. Its number of parameters (synapses) grows exponentially with the number of neurons and layers. To uncover topological properties emerging in such systems, this ample parameter space needs to be explored in more depth. However, it becomes impractical to compute every possible combination to even small architectures. Therefore, we propose constructing a neural network population with different properties within the most common practical scenarios in the field. This approach allows the comparison between a wide range of neural networks, which is one of the main differences to previous works such as \cite{testolin2020deep,zambra2020emergence}.

\subsection{Dataset construction: training a neural network population}

We propose a new neural network dataset composed of a wide range of synapse configurations. The idea is to use a fixed architecture (number of neurons and synapses) and training hyperparameters, then train networks of different initial weights. For that, we vary the seed of the random number generator that produces the synapse distributions (for specific details, check our repository \textsuperscript{\ref{github}}). In this sense, it is possible to widely explore the synapse space of random initial networks and analyze how it impacts the training dynamics and final behavior. After training, even networks of similar performance have different initial weight configuration, allowing us to analyze recurrent CN properties emerging between them. It is important to notice that the idea for building our dataset is different from previous works on Neural Architecture Search (NAS) \cite{ying2019NAS}. In these works, the goal is to search and optimize architectures for better performance, while here we are storing the entire network for tacking the properties which explain their performance.

\begin{figure}
    \centering
    
    \subfigure[Fully connected neural network.]{\includegraphics[width=0.45 \linewidth]{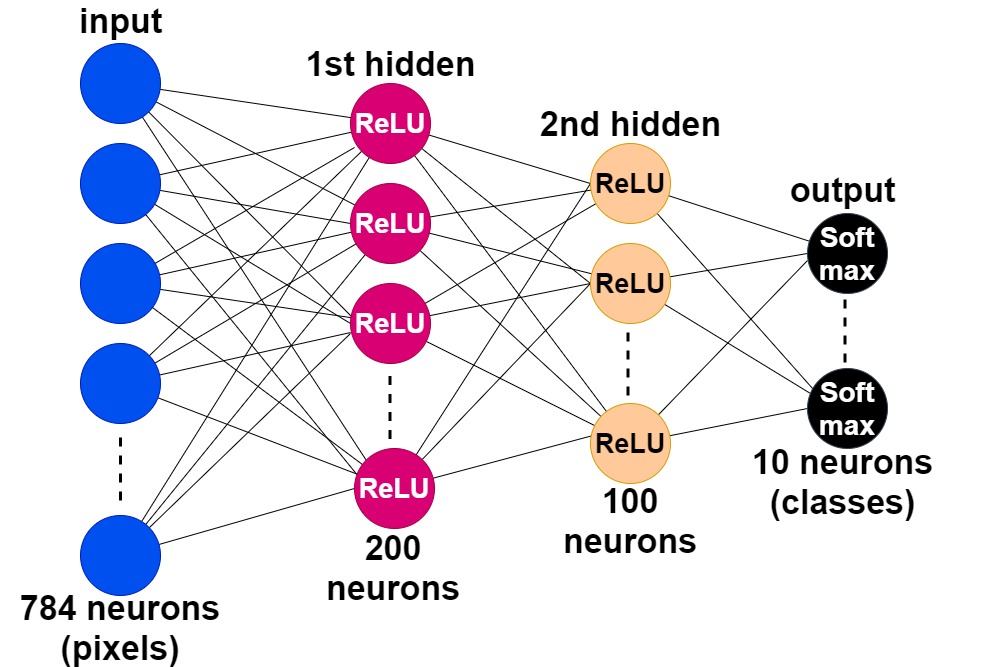}
    }
    \subfigure[Visualization of all its neurons and synapses.]{\includegraphics[width=0.45\linewidth]{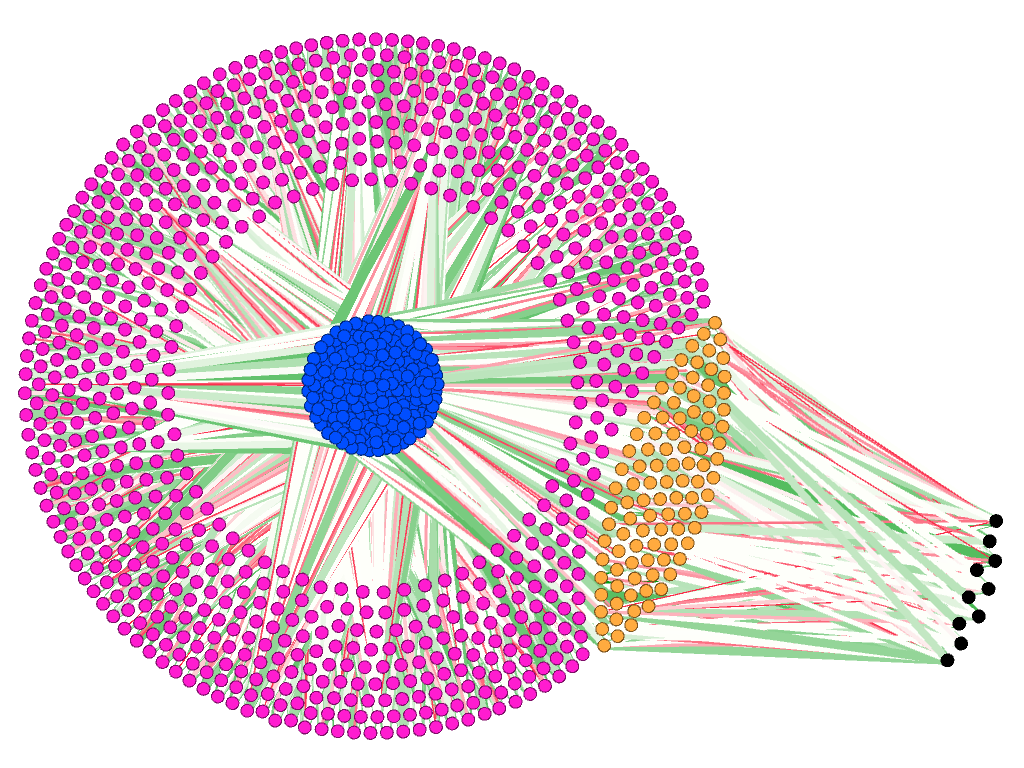}
    }
    
    \caption{\label{fig:deepbelief} The architecture used on this work with its corresponding configuration, and a visualization of all its neurons (1,094 nodes) and synapses (177,800 edges) as a weighted and undirected network.}
\end{figure}

Firstly we define the architecture for our study. This choice depends on the problem one needs to address; we focused on supervised classification on vision tasks. An MLP-like architecture is considered, where the input patterns are grayscale images of sizes 28x28, which are fed into a flattened layer of 784 neurons. Two hidden layers, with 200 and 100 neurons ($n=300$), respectively, employ the ReLU activation function. The output layer comprises ten neurons with a softmax activation function, representing a 10-class classification task (benchmarks are described later). Here we do not consider bias terms for simplicity. This architecture yields 177,800 trainable parameters. Figure \ref{fig:deepbelief} (a) shows a summary of the model, which we implemented using the Keras 2.2.4 library \footnote{\url{https://github.com/keras-team/keras/releases/tag/2.2.4}}. In Figure \ref{fig:deepbelief} (b), we show all the model's neurons and synapses using a network visualization technique that clusters vertices according to their degree. Neuron colors represent its corresponding layer, while edge colors refer to the synapse signal (green for positive and red for negative).

\begin{figure}
    \centering
    \newcommand\sampleimg{0.8}
    \subfigure[MNIST]{\includegraphics[width=\sampleimg \linewidth]{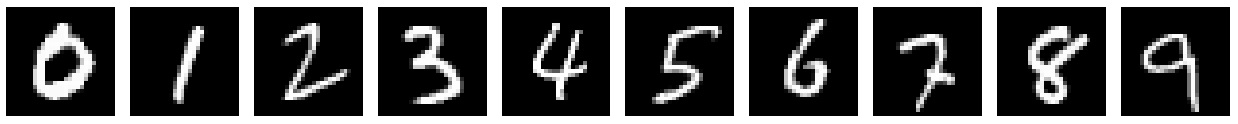}}
  
    \subfigure[Fashion MNIST]{\includegraphics[width=\sampleimg \linewidth]{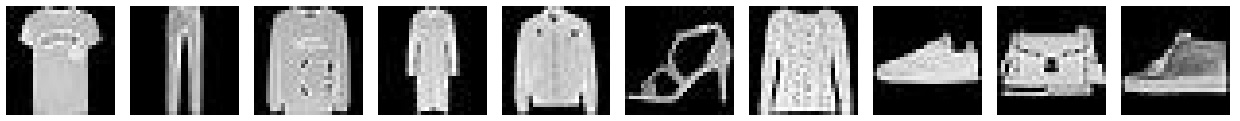}}
    
     \subfigure[CIFAR-10]{\includegraphics[width=\sampleimg \linewidth]{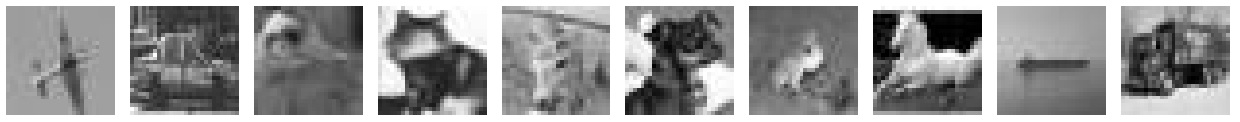}}

 \subfigure[KTH-TIPS]{\includegraphics[width=\sampleimg \linewidth]{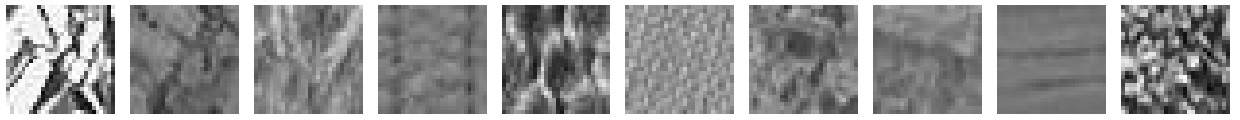}}

    \caption{\label{fig:datasets}Some samples from the four vision benchmarks where the fully connected neural networks are applied.}
    
\end{figure}

To achieve different networks from this same architecture, we vary the weight initialization. Novel techniques, such as the Glorot initializer \cite{glorot2010understanding}, optimize the weight distribution by fitting a normal curve around 0 with a small standard deviation. In practice, this tends to produce similar, more stable networks and may help with training. However, our focus here is not on better performance but on heterogeneity. Therefore, we use an initialization technique that allows a broader range of possible weight configurations. We consider a uniform distribution in the interval $[-0.9, 0.9]$. Using a wider range such as $[-1, 1]$ tends to produce poorer models, while compressing it tends to produce more similar models. The considered interval generates more diverse networks at both ends regarding performance in different vision benchmarks. For the training, we considered the following: SGD optimizer on a categorical cross-entropy loss function, learning rate $\alpha=0.01$, and batch size 100. We observed that the training usually stabilized around 15 to 20 epochs, then a total of 30 training epochs were considered. With the described architecture and training procedure, we address the following four vision benchmarks:
\begin{itemize}
    \item MNIST \cite{lecun1998gradient}: This is one of the most known computer vision benchmarks, with grayscale 28x28 images representing handwritten digits from 0 to 9; some samples are shown on Figure \ref{fig:datasets} (a). It comprises 60 thousand images for training and 10 thousand for the test where the goal is a 10-class classification problem to identify each digit. It is important to note that all the other four benchmarks described here have the same number of classes.

    \item Fashion MNIST \cite{xiao2017fashion}: This dataset consists of a more complex visual structure, with detailed objects rather than almost-binary pixels in comparison to MNIST. Fashion MNIST is composed of grayscale clothing images, with the following classes: t-shirt, trouser, pullover, dress, coat, sandals, shirt, sneaker, bag, and ankle boots. Samples of each class are shown on Figure \ref{fig:datasets} (b).

    \item CIFAR-10 \cite{cifar}: To include a different visual problem, we considered this object classification dataset. The CIFAR-10 variant comprises grayscale samples of 10 classes: airplane, automobile, bird, cat, deer, dog, frog, horse, ship, and truck. It is a more challenging classification problem as images are uncontrolled and objects conditions and positions vary greatly (some samples are shown in Figure \ref{fig:datasets} (c)).
    
    \item KTH-TIPS \cite{caputo2005class,mallikarjuna2006kth}: Texture classification is another important problem in computer vision. Therefore we considered this well-known benchmark. The original KTH-TIPS 2 dataset comprises RGB images of 10 texture classes: aluminum foil, cork, wool, corduroy, linen, cotton, brown bread, white bread, wood, and cracker. We built a different dataset by cropping the original 200x200 samples (different sizes are not considered) into 28x28 non-overlapping grayscale images. We divide the new crops into training and testing folds by ensuring that no crops of the same image appear together in the same fold. It yields a total of 165 thousand samples for training and 35 thousand for the test. We then randomly choose 60 thousand images for training and 10 thousand for the test. Our version of KTH-TIPS is available at \footnote{\url{http://scg-turing.ifsc.usp.br/data/repository.php}}, samples of each class are shown on Figure \ref{fig:datasets} (d).

\end{itemize}

\begin{figure}
    \centering

    \subfigure[MNIST, average train accuracy 80.6($\pm 12.5$) and test accuracy 78.8($\pm 12.1$).]{\includegraphics[width=0.4 \linewidth]{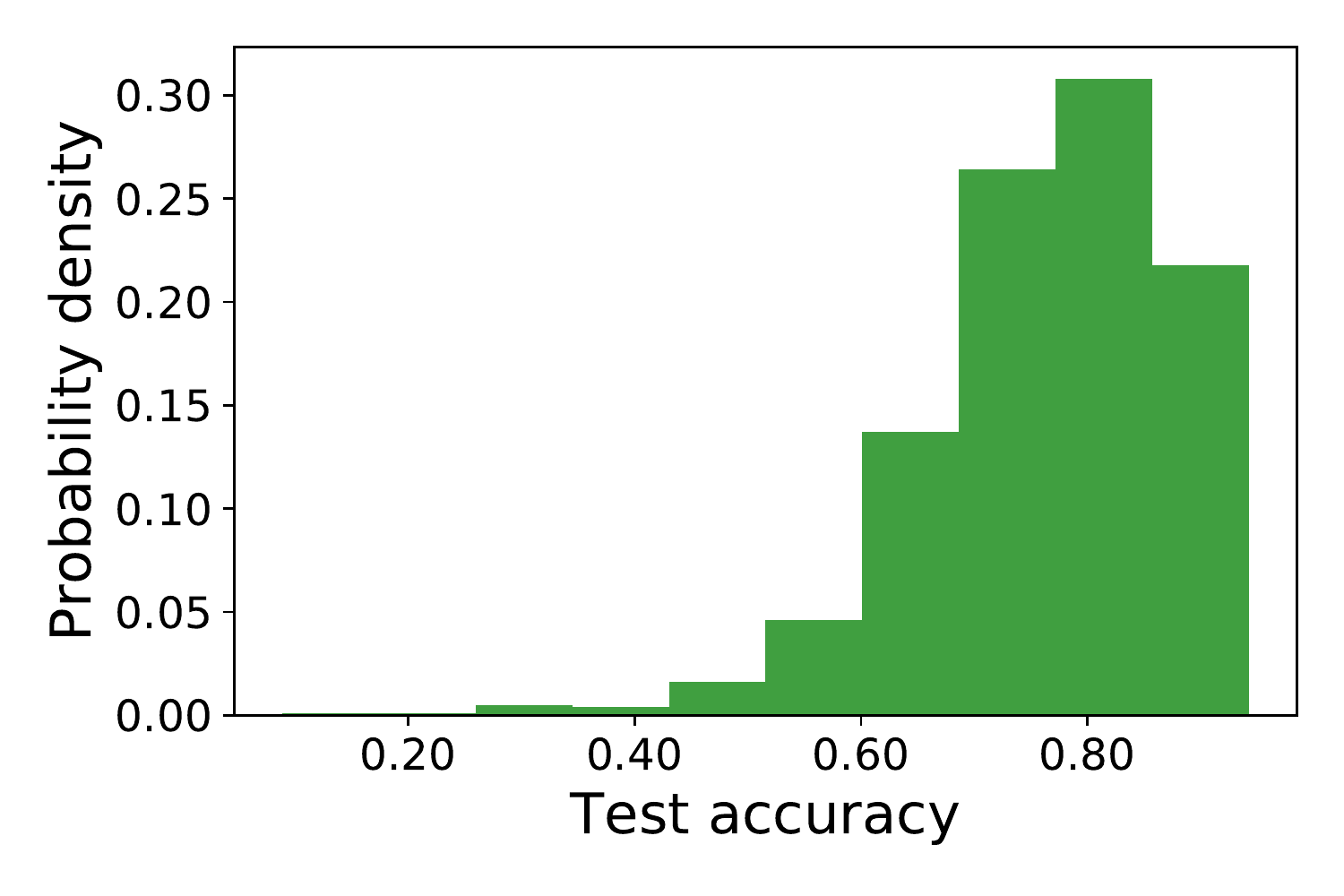}} \ \ \ \ \ \ \subfigure[Fashion MNIST, average train accuracy 62.2($\pm 16.9$) and test accuracy 60.7($\pm 16.3$).]{\includegraphics[width=0.4 \linewidth]{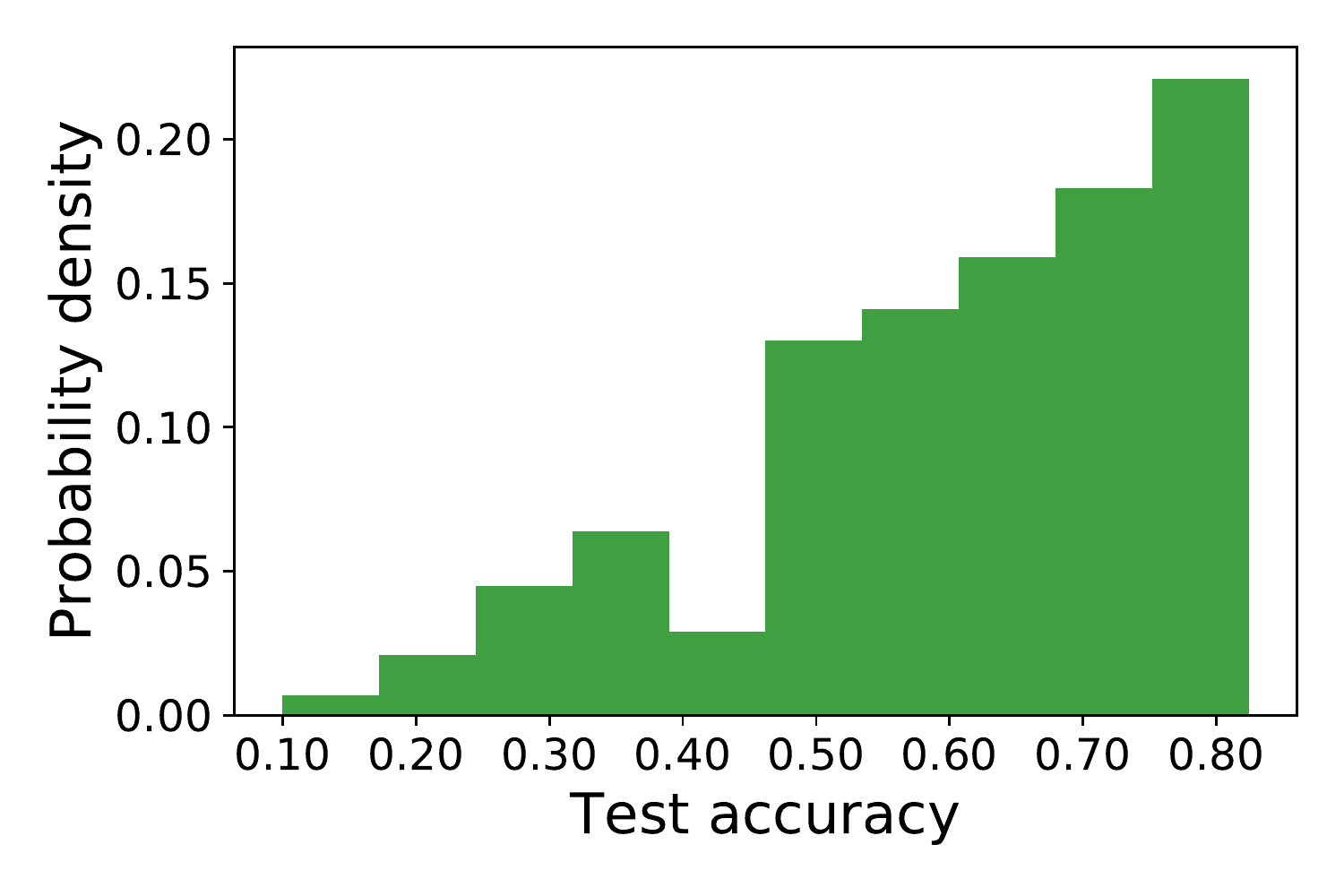}} \\ \subfigure[CIFAR-10, average train accuracy 23.8($\pm 2.8$) and test accuracy 23.4($\pm 2.7$).]{\includegraphics[width=0.4 \linewidth]{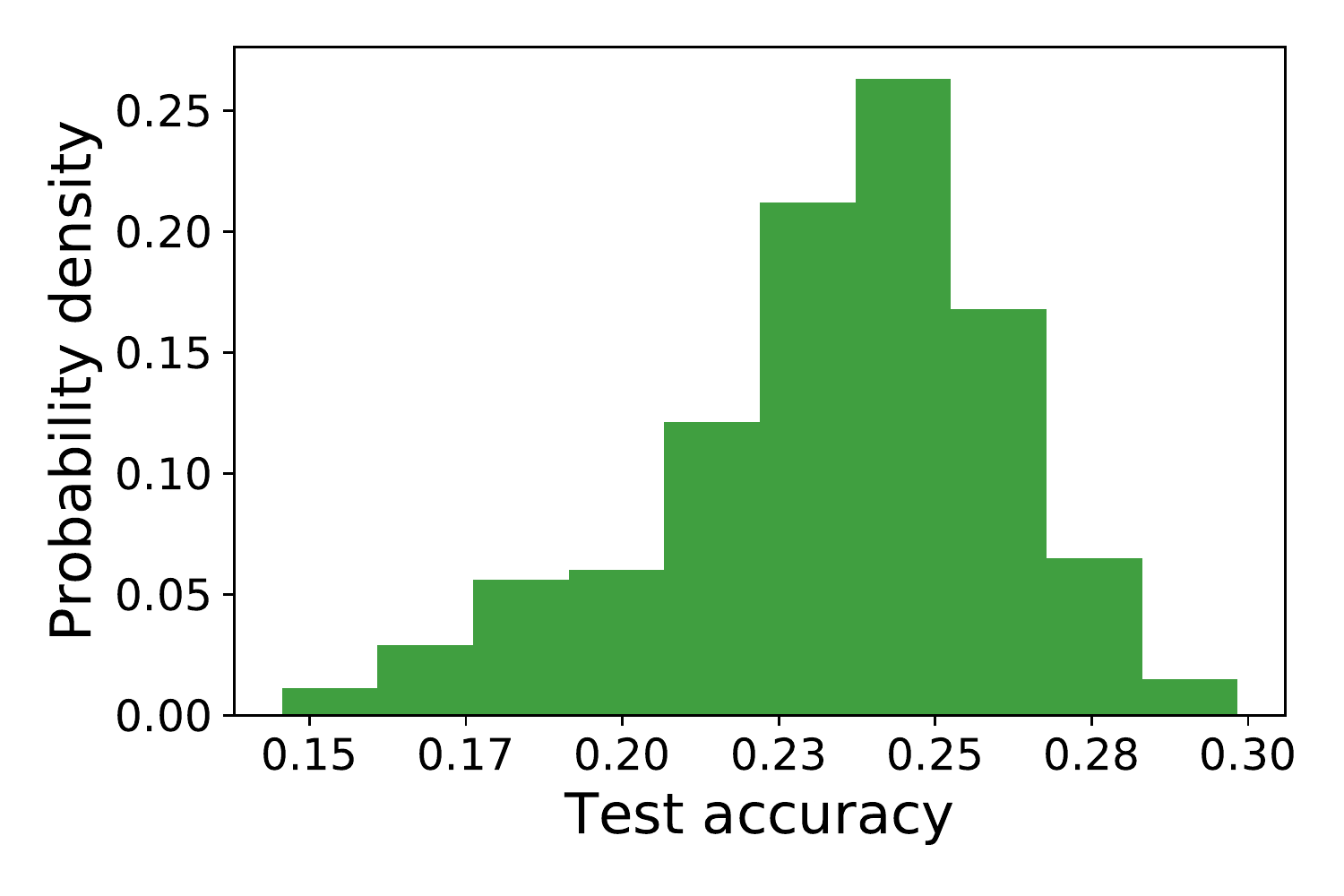}} \ \ \ \ \ \ \subfigure[KTH-TIPS, average train accuracy 14.8($\pm 2.4$) and test accuracy 14.6($\pm 2.4$).]{\includegraphics[width=0.4 \linewidth]{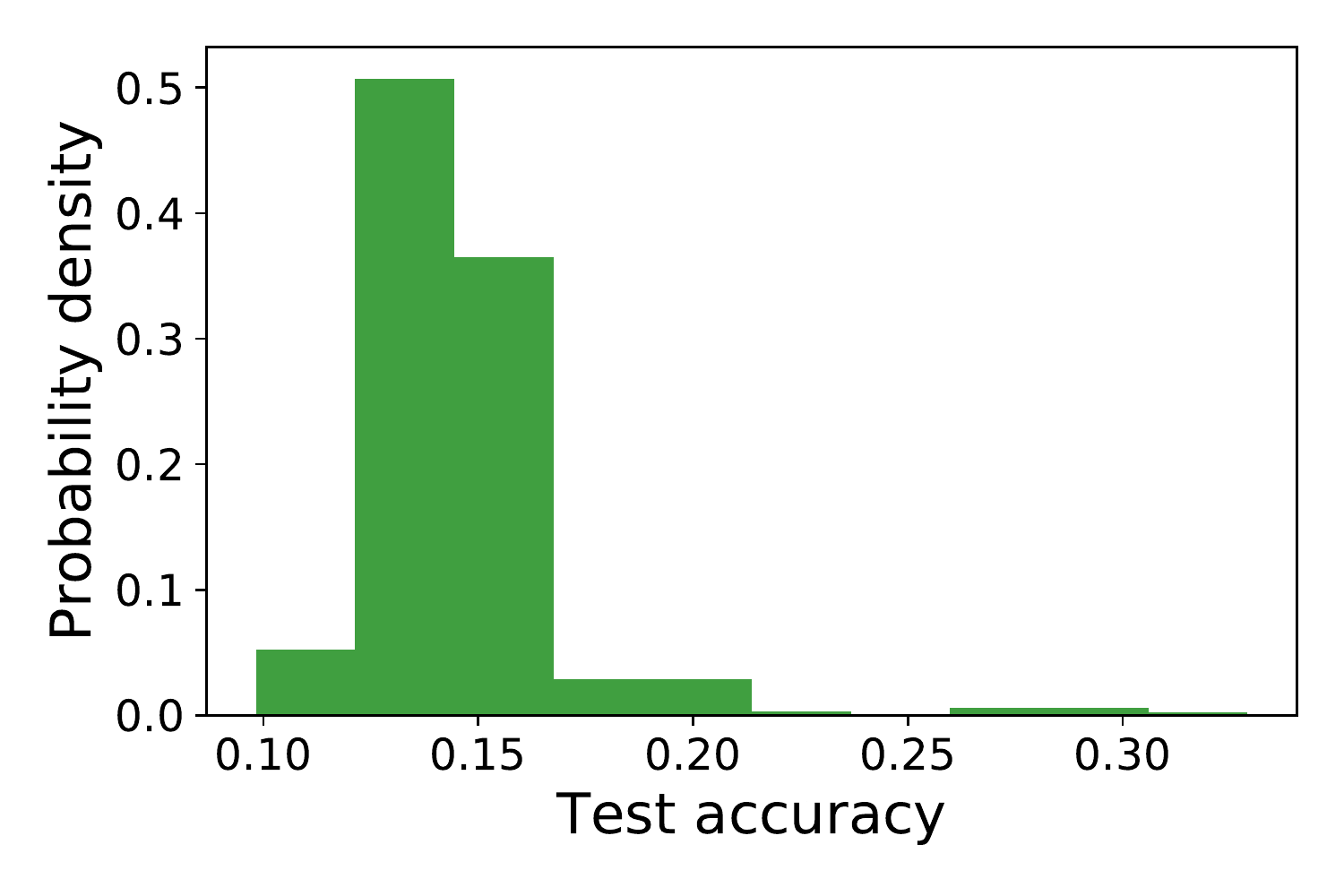}}
    
    \caption{\label{fig:dataset_accuracy}Test accuracy distribution of the proposed neural network dataset. It contains 1000 samples of fully connected and 4-layers deep neural networks trained on each benchmark, totaling 4000 samples.}
    
\end{figure}

For each dataset, we train a population of 1000 neural networks, each one with a different random seed for weight initialization. The remaining configuration, such as training algorithms and batch order, are kept fixed for all networks. The result is a fairly distributed population of 4000 models in terms of final test accuracy, as Figure \ref{fig:dataset_accuracy} shows.

\subsection{Neuronal Complex Network characterization}

Considering the neural network population, we compute CN centrality measures from its neurons. Each network can be directed modeled as weighted, undirected, and bipartite graphs. Consider a neural network $\aleph=\{L_0,...,L_d\}$ with depth $|\aleph| = d+1$ (number of layers), where $L_i=[n_1,...,n_l]$ represents its $i$th layer with $l$ neurons and a weight matrix $W_i$, while $L_0$ and $L_d$ are its input and output layers, respectively. We then construct a weighted graph $G(\aleph) = (V, E)$, where $V=[L_i]$ is its set of vertices representing all neurons, and $E=[W_i]$ their synapses for $i \in [0,d]$. The graph structure is built upon neurons and synapses and does not consider the activation function or the direction of connections.

Given the new graph representation $G$, centrality measures can be computed to describe each neuron $n_i$ ($f(n_i)$). This is a classic approach in CN research, where methods try to estimate the vertex centrality, or importance, for the model structure and dynamics. Vertex centrality is, therefore, an important property of CNs and have helped to identify critical properties in a wide range of real-world systems \cite{aplicacoesRC}. Considering that the literature is rich on such methods \cite{cnusp}, it is possible then to characterize neurons in different perspectives, as different measures has a different topological meaning. For that, we compute centrality from hidden neurons (layers $[L_1,...,L_{d-1}]$), i.e., as our samples are 4-layers deep, only neurons from layers $L_1$ and $L_2$ are considered. The input and output layers are not considered as they are structurally different (connections in just one direction), and the first and last hidden layers already include their connections. Considering a hidden neuron $n_i$ in the graph representation, it is possible to calculate a given measure $f(n_i)$ to characterize it. We narrowed down a set of specific measures considering different approaches/paradigms and some practical characteristics such as complexity/computation time and availability in known libraries (for easy reproduction). This was carried out considering the possibility of future practical applications of our findings, considering that some CN methods are too expensive to be employed in modern deep neural networks. It is also important to notice that some measures cannot provide useful information for regular/fully connected graphs. Therefore we employ some threshold techniques (described below).

The NetworkX 2.4 library \footnote{\url{https://networkx.org/documentation/networkx-2.4/}} is employed for computing network properties, which is one of the most complete and diffused frameworks in python \cite{hagberg2008exploring}. An analysis on the computational complexity and specific parameter details of each CN measure is beyond the scope of this work. The reader may refer to their original papers and also the NetworkX documentation, which considers the best practices in terms of implementation and analysis of these methods. NetworkX and python are used also considering another important properties such as open source code and the ease to combine it with the most common neural network libraries (Tensorflow and Keras), allowing us to release our implementation in a single programming language \textsuperscript{\ref{github}}. Finally, we describe the CN measures in the following:

\begin{itemize}
    \item \textbf{Weighted degree, or strength} $\mathbf{(s)}$ Equation \ref{eq:s}: Represents the neuron strength, the rate at which it lets information flow. High positive strength means an excitatory neuron that gives higher importance to both its incoming and outgoing connections. Near zero means average neuron, a neutral neuron (connections near zero), or a neuron with half positive and half negative connections, configuring an antagonistic neuron. Negative strength is the predominance of negative connections, which means inhibitory neurons. This measure is computed over the original neural network, with all its edges and weights.

    \item \textbf{Average neighbor strength} $\mathbf{(s^w_{nn})}$ Equation \ref{eq:sw}: Similar to the neuron strength, but focuses on the first-order neuron neighbors. In a neural network, neuron neighbors are the neurons on the previous and the next layer. It means that the strength of $i$ depends on the strength of its neighbors. Therefore, if they have high strength, $i$ probably also have high strength. This measure is computed over the original neural network, with all its edges and weights.
    
    \item \textbf{Second order centrality} $\mathbf{(so)}$ Equation \ref{eq:so}: This measure is based on the concept of random walks, and is computed over a thresholded version of the original neural network, where we keep only positive connections. It represents the standard deviation of the return times to a given neuron of a perpetual random walk on the neural network.
    
    \item \textbf{Subgraph centrality} $\mathbf{(sg)}$ Equation \ref{eq:sg}: Represents the sum of weighted closed walks of all lengths starting and ending at a given neuron. A closed walk is associated with a connected subgraph, i.e., a subset of interconnected neurons within the ANN. This measure is computed using a spectral decomposition of the adjacency matrix over a thresholded version of the original neural network, where we keep only positive connections.
    
    \item \textbf{Number of maximum cliques} $\mathbf{(mc)}$ Equation \ref{eq:mc}: In a neural network, a clique is a subset of neurons such that every two neurons are adjacent. Considering that the unweighted neural network is regular, a thresholded version is used (only positive connections). In this scenario, a maximum clique is a path between excitatory synapses joining the largest number of neurons. Therefore, this measure counts the number of such structures in which a neuron participates, representing its importance for input stimulation.

    \item \textbf{Bipartite local clustering} $\mathbf{(bc)}$ Equation \ref{eq:bc}: As the neural network is a multipartite graph, we consider this measure, which estimates the local neuron density according to second-order neighbors. The traditional clustering coefficient cannot be computed for neural networks, as there are no triangles between neurons. This version then relaxes the original clustering coefficient for such structures. This measure is calculated over a thresholded and unweighted version of the original neural network, where we keep only positive connections.
    
    \item \textbf{Harmonic centrality} $\mathbf{(hc)}$ Equation \ref{eq:hc}: A path-based measure of centrality, which represents the sum of the reciprocal of the shortest path distances from all other neurons to a given neuron. This measure is computed over a thresholded version of the original neural network, where we keep only positive connections. 
    
    \item\textbf{Current flow closeness} $\mathbf{(CF_c)}$ Equation \ref{eq:cfc}: This is a path-based approach, inspired by the electric flow on circuits. Each neuron pair is considered a source-target pair, where current is injected at the source and drained at the target. The measure then quantifies the importance of neurons for the electric flow throughout all the neural network paths, i.e., the synapse flow. This approach is more realistic in a neural network than the other centrality measures which assume that information flows exclusively through the shortest paths. This measure is computed over the original neural network, with all its edges and weights.

\end{itemize}

Considering these centrality measures, we propose local neuron descriptors, or global average features from each hidden layer. Given a hidden layer $L_i$ with $n$ neurons, and a given measure $f$, a feature vector can be obtained $\upsilon_f (L_i)= [f(v_1),...,f(v_n)], \ \forall \ v \in L_i$. The vector $\upsilon_f$ preserves neuron order on the layer. However, this order is not meaningful as the network is fully connected, and it also depends on the random initialization. To remove this spatial information we compute the layer average
\begin{equation}\label{eq:layerdescriptor}
   \upsilon_{L_i,f} = \frac{1}{n} \sum \upsilon_f (L_i).
\end{equation}

For local characterization, it is possible to compute a set of descriptors to a given neuron by combining different measures. A local feature vector is then expressed as
\begin{equation}\label{eq:localdescriptor}
   \psi(v_i) = [f_j(v_i)]
\end{equation}
where $n_i \in V $ is a vertex representing a neuron from hidden layers, and $f_j$ is a CN measure ($f_j$ $\in$ $[s,$ $s^w_{nn},$ $CF_c,$ $bc,$ $sg,$ $hc,$ $so,$ $mc]$). In this sense, each neuron is a sample to be characterized and not the whole layer or neural network. By analyzing the local descriptors $\psi(v_i)$ from all neurons, it is then possible to characterize the local topological properties of the neural network.

We employ a non-supervised approach to find neuronal signatures by grouping local descriptors with similar features. This approach has been employed for image recognition and is usually called bag-of-visual-words \cite{csurka2004visual}. It derives from the bag-of-words method, which considers the word frequency (features) from a given vocabulary of $k$ types. In the case of visual words, features are real-valued vectors representing local image properties. Here, our features are neuronal topological signatures expressed through the CN measures of each neuron. Given a set of neural network graphs $P=[G_a]$, a matrix is built by stacking neuron descriptors from all neural networks in $P$
\begin{equation}\label{eq:D}
D=[\psi(v_i)], \ \forall \ v_i \in G_a, \ \forall \ G_a \in P    
\end{equation}
where $D$ has dimensions $|P|n_a$-by-$m$, where $|P|$ is the number of graphs in $P$, $n_a$ the number of hidden neurons in each neural network and $m$ the number of measures. We consider $P$ as the set of neural networks trained for a given vision benchmark, $n_a=n=300$ (see the architecture description), and $m$ is the number of considered CN measures (further discussed on Section \ref{sec:results}).

 $D$ represents all the CN neuronal properties from a set of networks $P$, which is considered to build the vocabulary of the $k$ most relevant features, or neuron groups. We call this method bag-of-neurons (BoN), where each vocabulary element represents a neuron type that usually occurs in these neural networks. To find them, the $k$-means clustering (or Lloyd's algorithm) is applied to obtain group centers in a non-supervised fashion. It finds $k$ centroids (group centers) by first starting with random positions and then iteratively averaging its distance to nearest sample points ($D$ rows). In other words, it minimizes the within-cluster sum-of-squares (inertia) and tends to converge to regions where samples cluster together. This algorithm is usually employed due to its simplicity and performance, and it is commonly cited among the best data mining methods \cite{wu2008top}. The $k++$ method \cite{arthur2006k} is considered for initialization (rather than regular random sampling), which spreads the initial clusters with probability proportional to its squared distance from the point's closest existing cluster center. We also consider relative tolerance $10^{-3}$ with regards to the Frobenius norm of the difference in the cluster centers of two consecutive iterations to declare convergence. The matrix $D$ is feature-wise normalized by the largest value to avoid disproportion in the Euclidean distance calculation. A total of 100 repetitions is performed each time we run the $k$-means algorithm, and the best result based on inertia is used. This result is a set of $m$-dimensional centroids $C=[\psi_{1},...,\psi_{k}]$ representing the features of the $k$ most common neuron types. It is important to notice that even when using $k++$ and performing 100 repetitions, the position of the centroids may still slightly vary by some decimal places, as the process is still random.

\section{Analysis and Results}\label{sec:results}

\begin{figure}[!htb]
\newcommand\resultimg{0.32}
    \centering
    \includegraphics[width=\resultimg \linewidth]{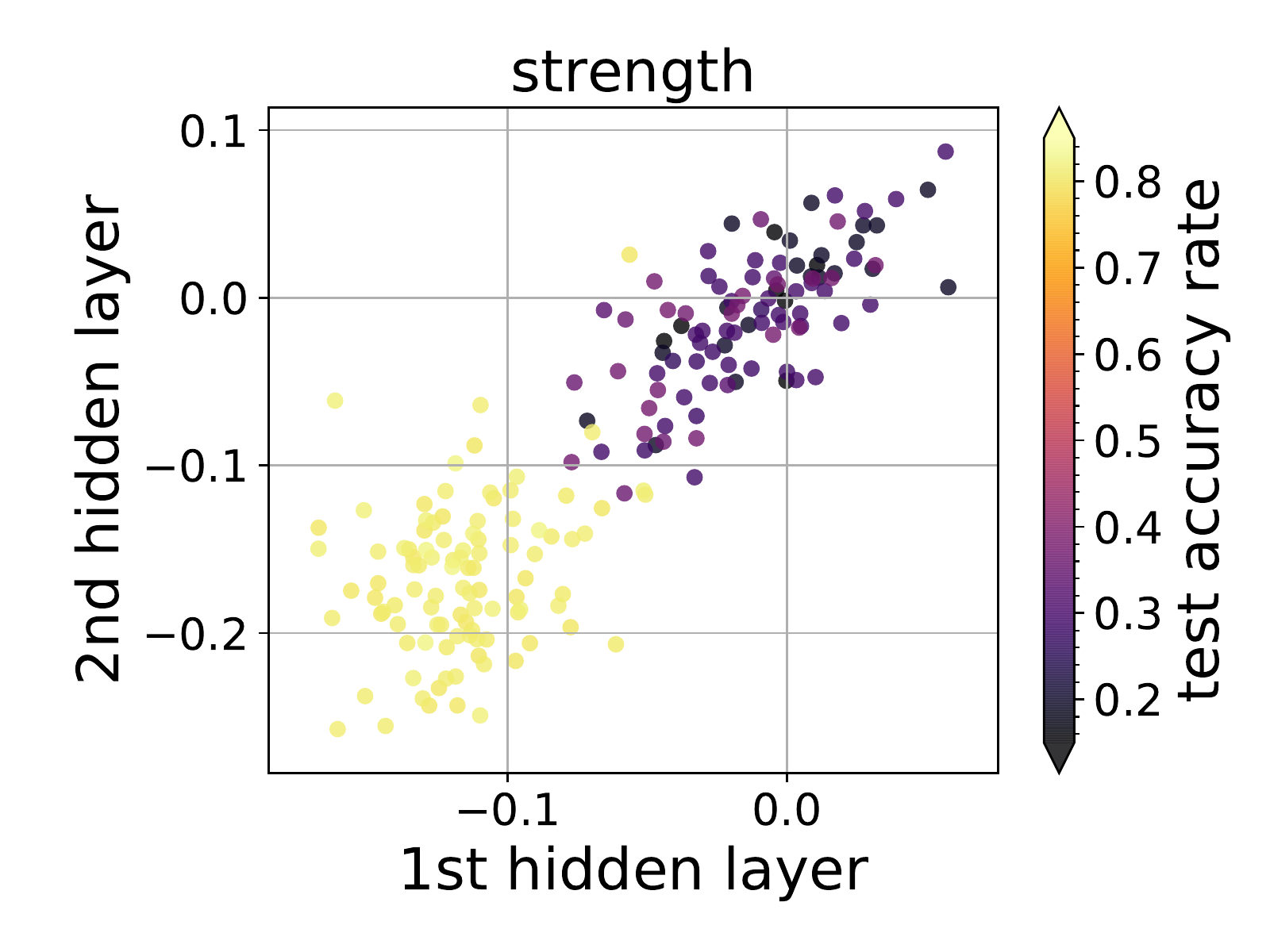} \includegraphics[width=\resultimg \linewidth]{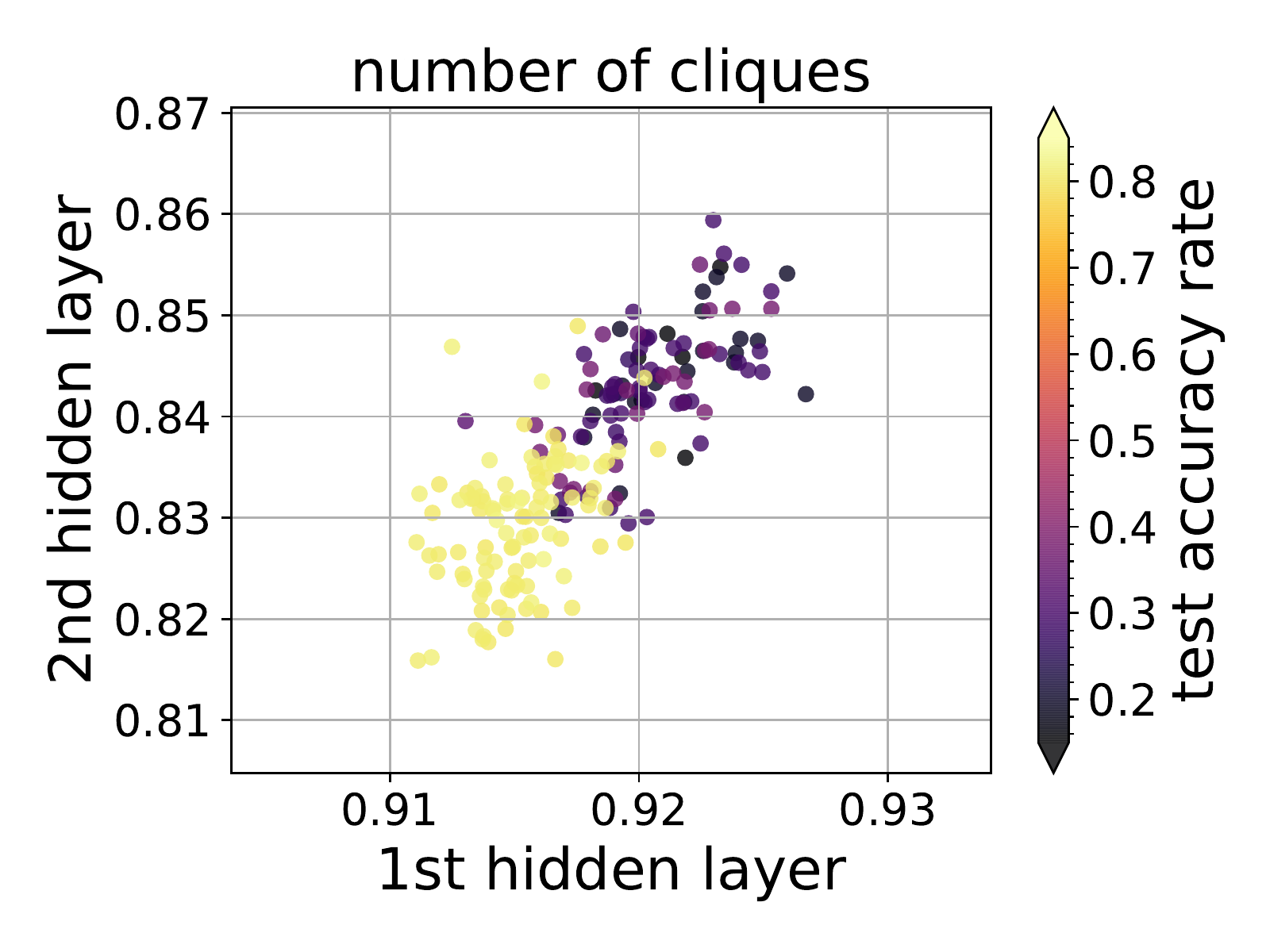} 
    \includegraphics[width=\resultimg \linewidth]{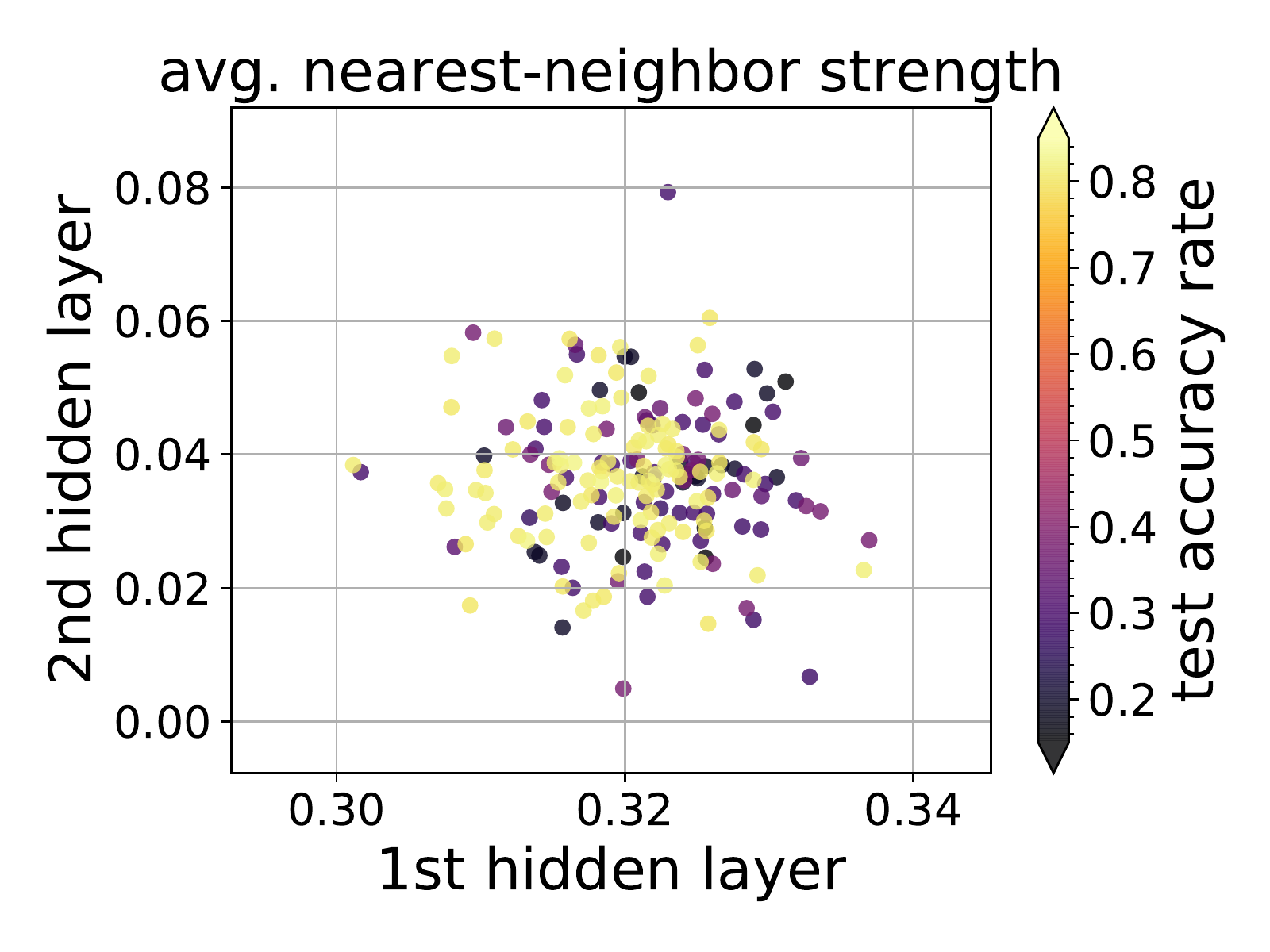}
    \\
    \includegraphics[width=\resultimg \linewidth]{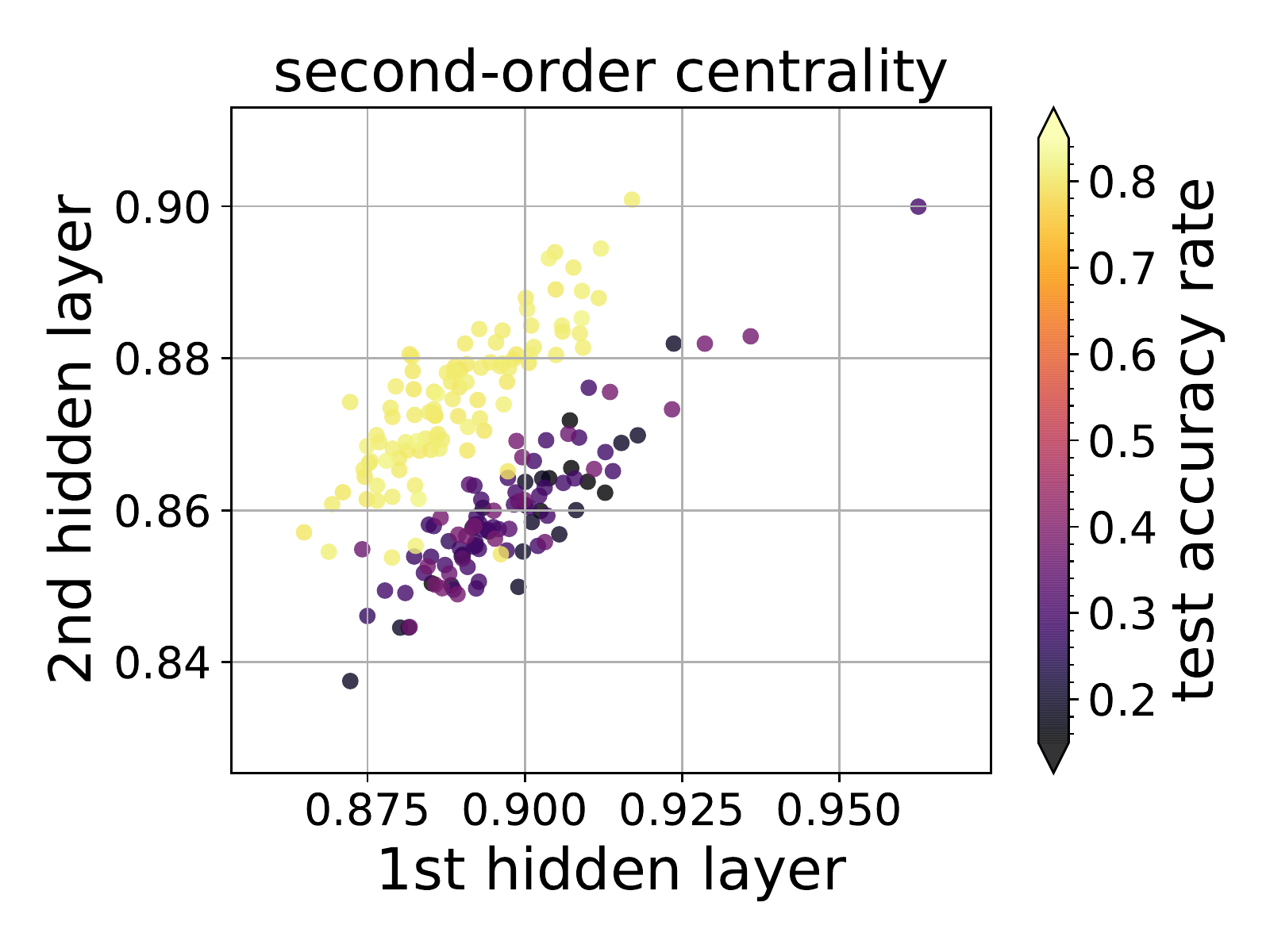}
    \includegraphics[width=\resultimg \linewidth]{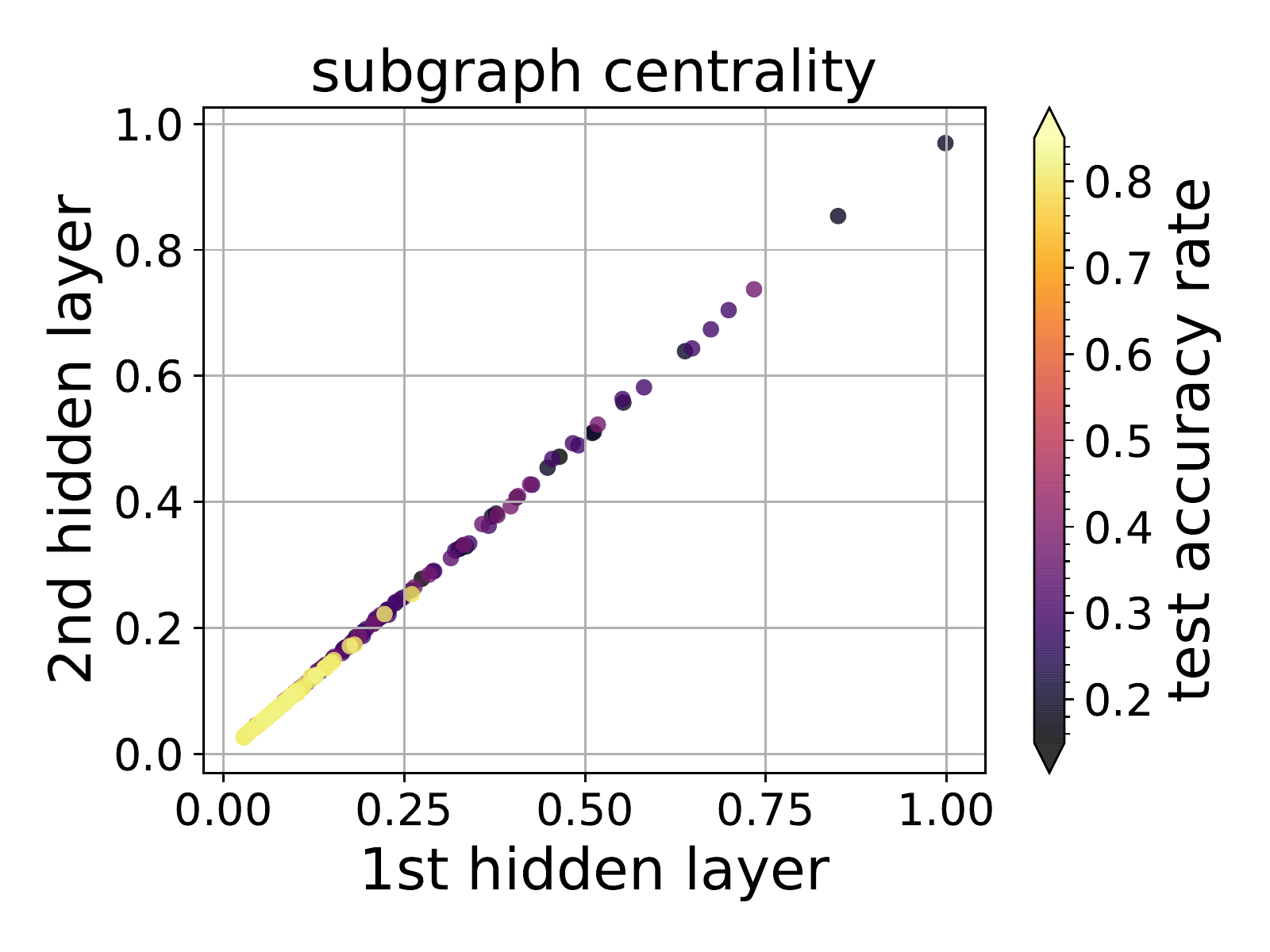} \includegraphics[width=\resultimg \linewidth]{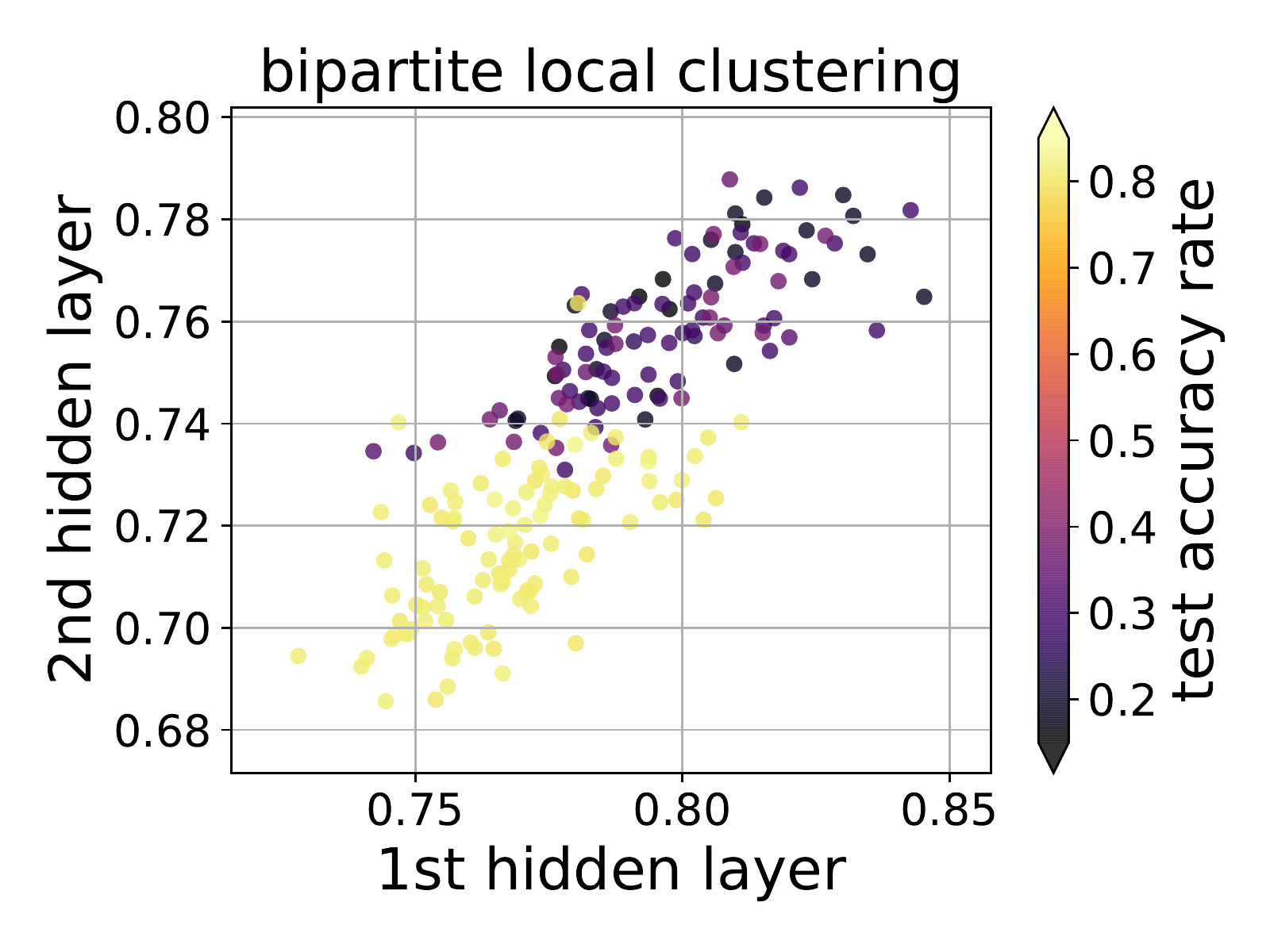}
    \\
    \includegraphics[width=\resultimg \linewidth]{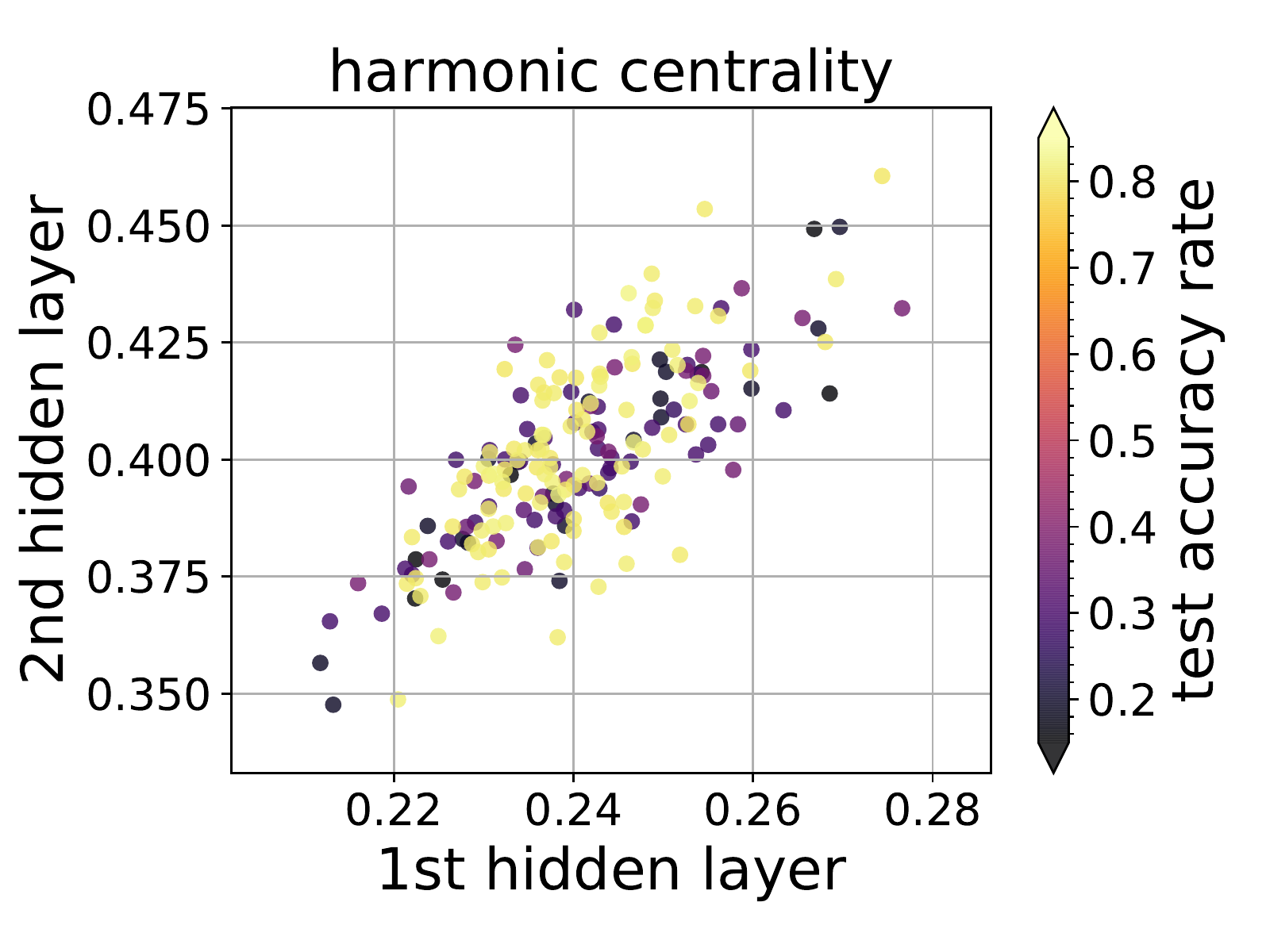} 
    \includegraphics[width=\resultimg \linewidth]{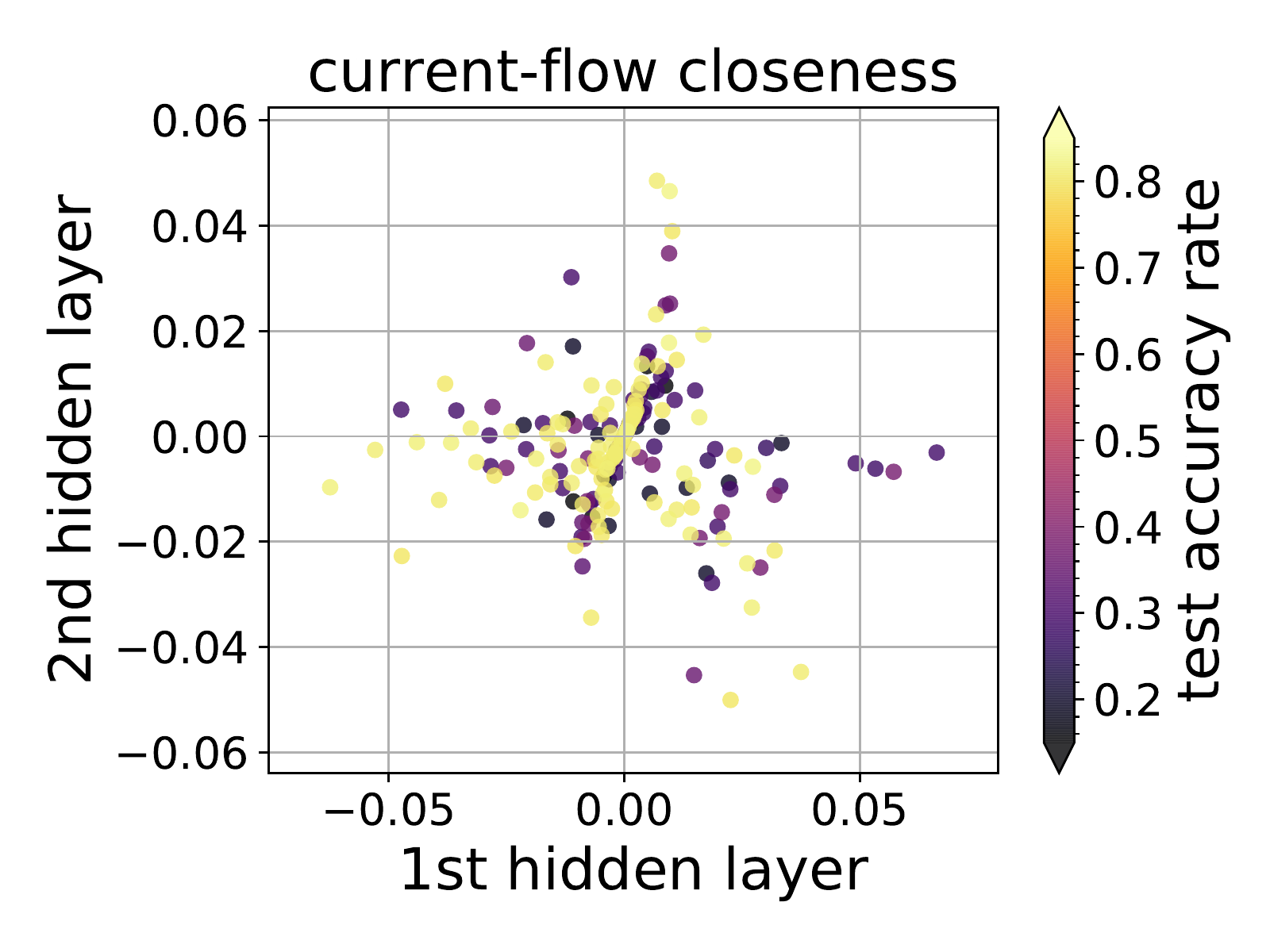} 
    
    \caption{\label{fig:single_measures_fminist}Spatial distribution of trained neural networks according to the average of their hidden neuron's topological measures ($\upsilon_{L_i,f}$). We show the best and worst 100 samples, and colors represent their respective final test accuracy considering the Fashion MNIST dataset.}
\end{figure}

We first analyze the apparent correlation between CN measures and the neural network performance by computing vectors $\upsilon_{L_i,f}$ for each hidden layer, given a measure $f$ (see Equation \ref{eq:layerdescriptor}). It summarises all the neural information into two values (1 for each hidden layer) for each measure. For a qualitative analysis, the information is visualized in two axis where $x = \upsilon_{L_1,f}$ and $y = \upsilon_{L_2,f}$, representing the 2D spatial distribution of each neural network. Results are shown in Figure \ref{fig:single_measures_fminist} and contain the best, and worst 100 samples in terms of test accuracy on the Fashion MNIST dataset, and color represents the test accuracy. It is possible to notice that some measures projections are practically linearly separable considering the two performance groups of networks, more specifically the strength ($s$), bipartite local clustering ($bc$), second-order centrality ($so$), and the number of cliques ($mc$). The subgraph centrality ($sg$) shows a total correlation between the first and second hidden layer, with a small difference in scale probably related to the different number of neurons and synapses it contains. Nonetheless, it is still possible to distinguish the two groups of networks even if considering the information from a single axis.

On the other hand, results suggest that the average nearest-neighbor strength ($s^w_{nn}$), current-flow closeness ($cfc$), and harmonic centrality ($hc$) are not able to express the correlation between neural network topology and performance. It is not easy to directly determine why they fail to tackle the correlation, as they represent different aspects from the graph. For instance, in \cite{testolin2020deep} authors have shown that the average nearest-neighbor degree may be correlated to the functioning of some receptive fields, but they do not discuss it in depth. However, it is not intuitive to believe that the assortativity principle take place on fully connected feedforward neural networks in terms of CN properties. For instance, considering neurons from the first hidden layer its neighbors are, respectively, the input and second hidden-layer, which most probably have a different role in the network dynamics. The same accounts for the neighbors of neurons from the second hidden layer, i.e., the output and first hidden layers. Therefore, considering the 4-layer architecture we analyze, our results suggests the absence of correlation between assortativity and performance. On the other hand, this assumption may not hold for deeper networks and broader vision tasks, as larger models may contain more redundancy.

Results also suggests that the CF closeness ($cfc$) and harmonic centrality ($hc$) are not directly related to neural network performance. Considering that both are path-based measures, we conclude that this kind of information may not correlate directly to the model accuracy. Therefore, as our goal on this work is to understand structure and performance, they are not considered for the next analyses (along with $s^w_{nn}$). Nonetheless, they are present on the dataset we make available \textsuperscript{\ref{github}} and further studies on different scenarios would be interesting.

\begin{figure*}[!htb]
    \centering
    \newcommand\tinyimg{0.857}
     \subfigure[First hidden layer.]{\includegraphics[width=\tinyimg\linewidth]{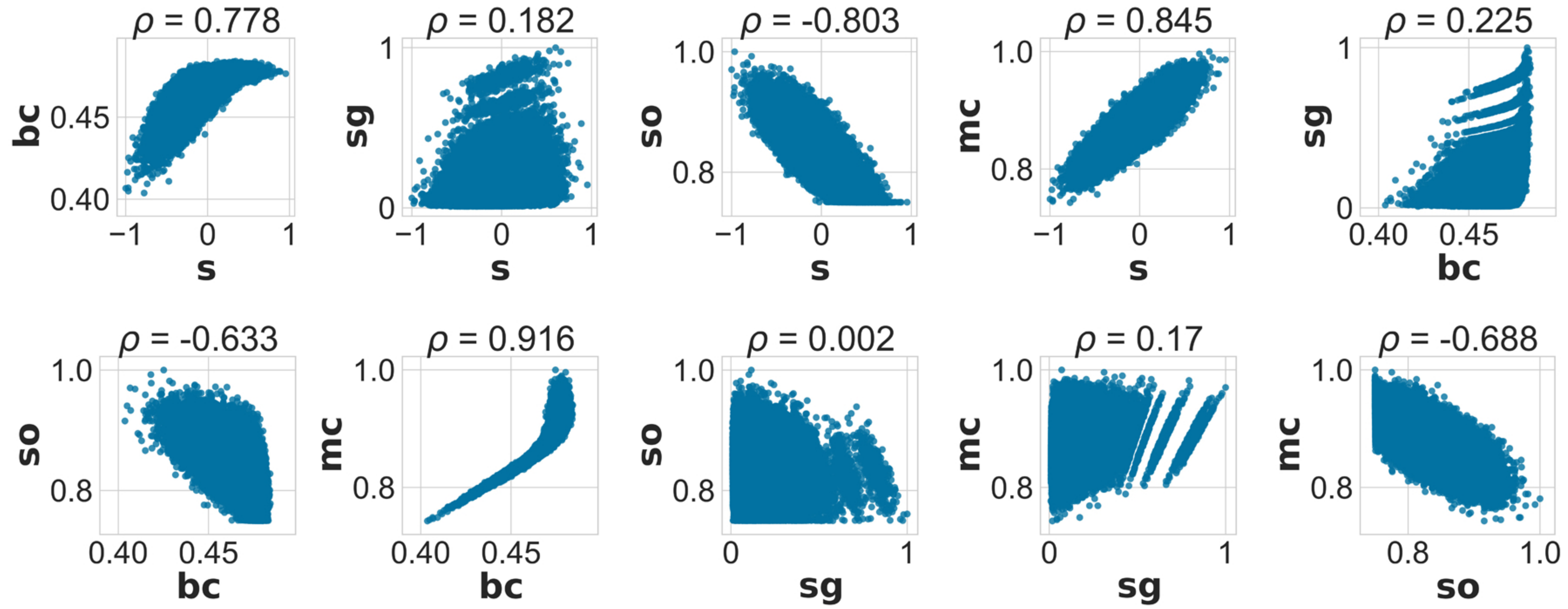}}
    
     \subfigure[Second hidden layer.]{\includegraphics[width=\tinyimg\linewidth]{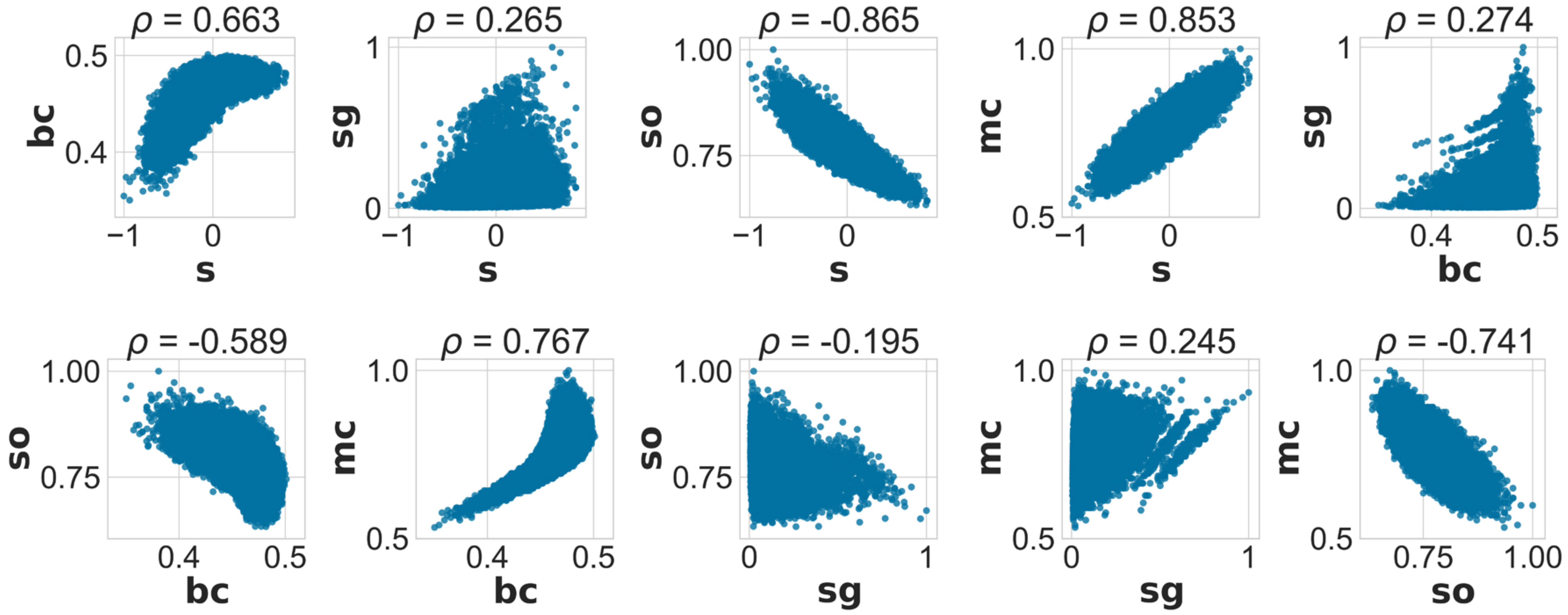}}

    \caption{\label{fig:measures_correlation} Correlation between pairs of different CN measures ($f_j(v_i)$) from hidden neurons of networks trained on Fashion MNIST. Considering the first hidden layer, there are 200 thousand neurons, and considering the second hidden layer, there are 100 thousand. $\rho$ is the Pearson's correlation between measure pairs. }
    
\end{figure*}

Different centrality measures may have considerable correlation even in networks from diverse domains \cite{meghanathan2015correlation}. For instance, random walks may be attracted to higher degree nodes, reflecting the degree distribution. Therefore, we consider the Pearson's correlation between measure pairs to discard redundant information. The local descriptor $\psi(v_i)$ of each neuron (see Equation \ref{eq:localdescriptor}) is computed considering each measure individually, resulting in a single value. We then cross these values for each measure pair, for each hidden layer separately. Each plot on Figure \ref{fig:measures_correlation} shows the measure pairs for all neurons from the 1000 neural networks (trained on Fashion MNIST), along with the corresponding Pearson's correlation $\rho$. If a pair of measures have $|\rho| > 0.8$, we remove one of them to reduce redundancy and keep the simplest measure (lowest computer cost). The strength has a high correlation with the second-order centrality and maximum cliques; thus, we keep the former and remove the other two. The removed measures ($so$ and $mc$)  also present a considerable correlation between themselves, corroborating that they provide low contribution for characterizing the neural network performance.

\subsection{Bag-of-Neurons properties}

\begin{figure}[!htb]
    \centering
    \subfigure[Fashion MNIST, individual hidden layers.]{\includegraphics[width=0.4 \linewidth]{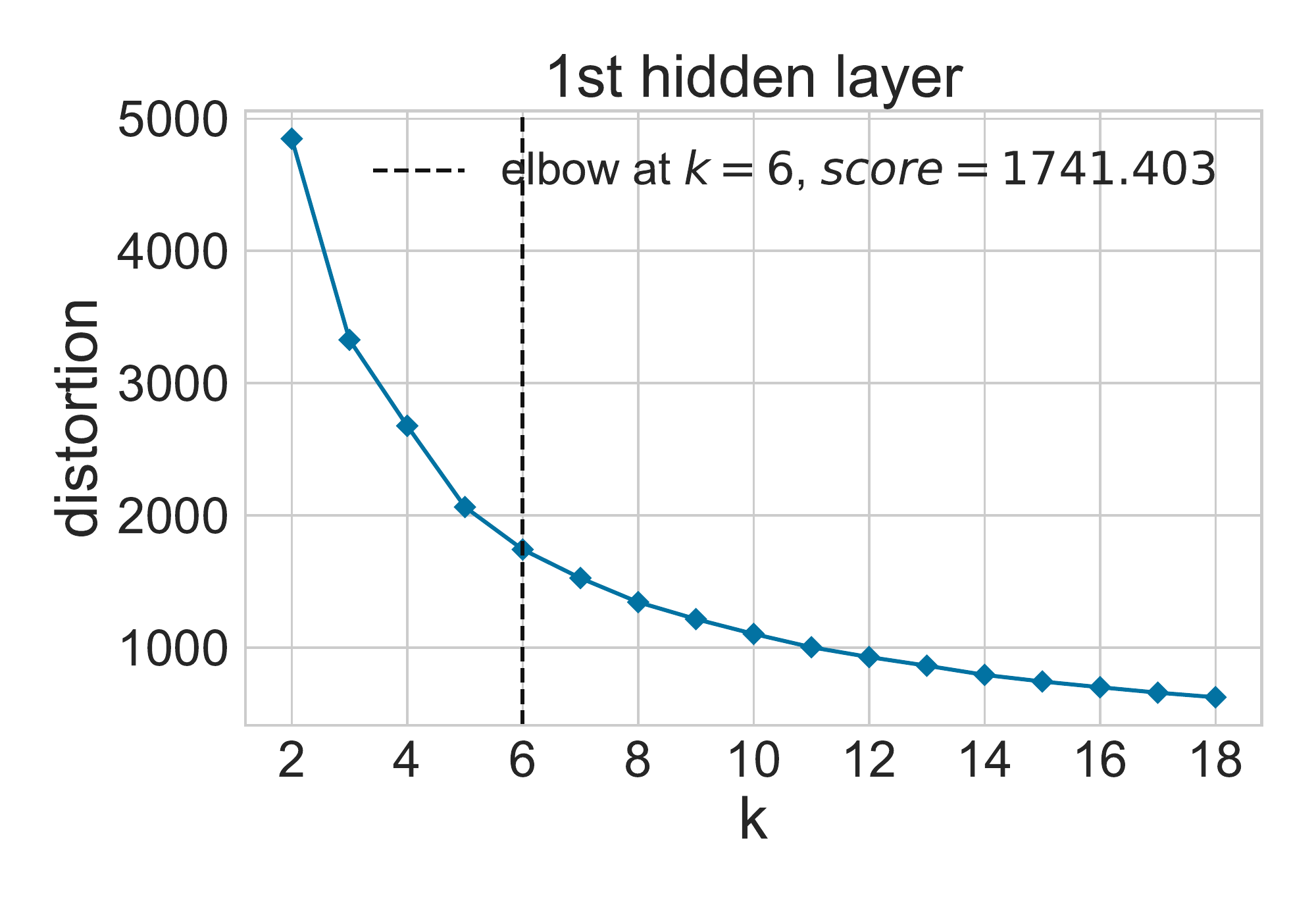} \includegraphics[width=0.4 \linewidth]{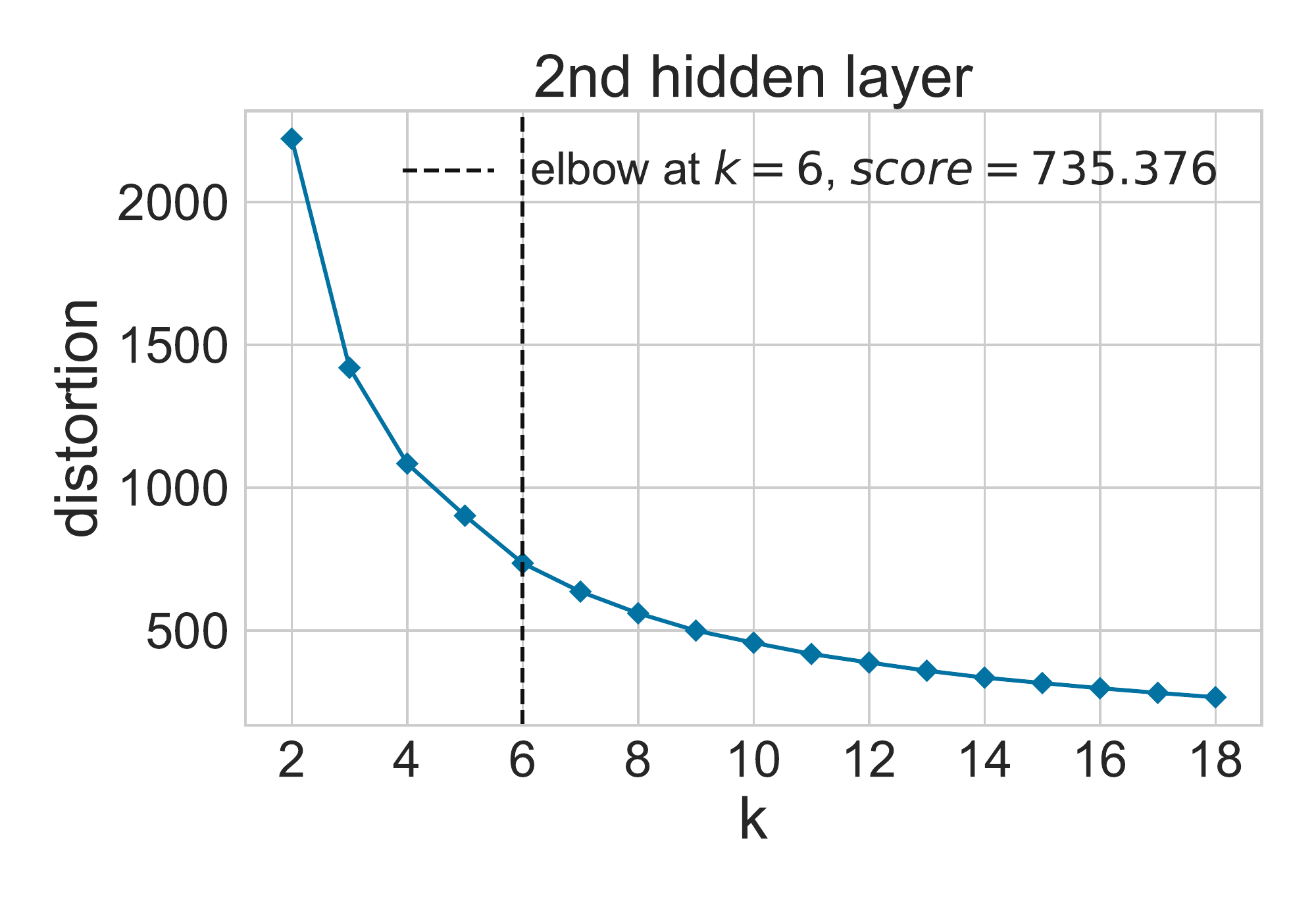}}
      \\
      \subfigure[MNIST.]{\includegraphics[width=0.4 \linewidth]{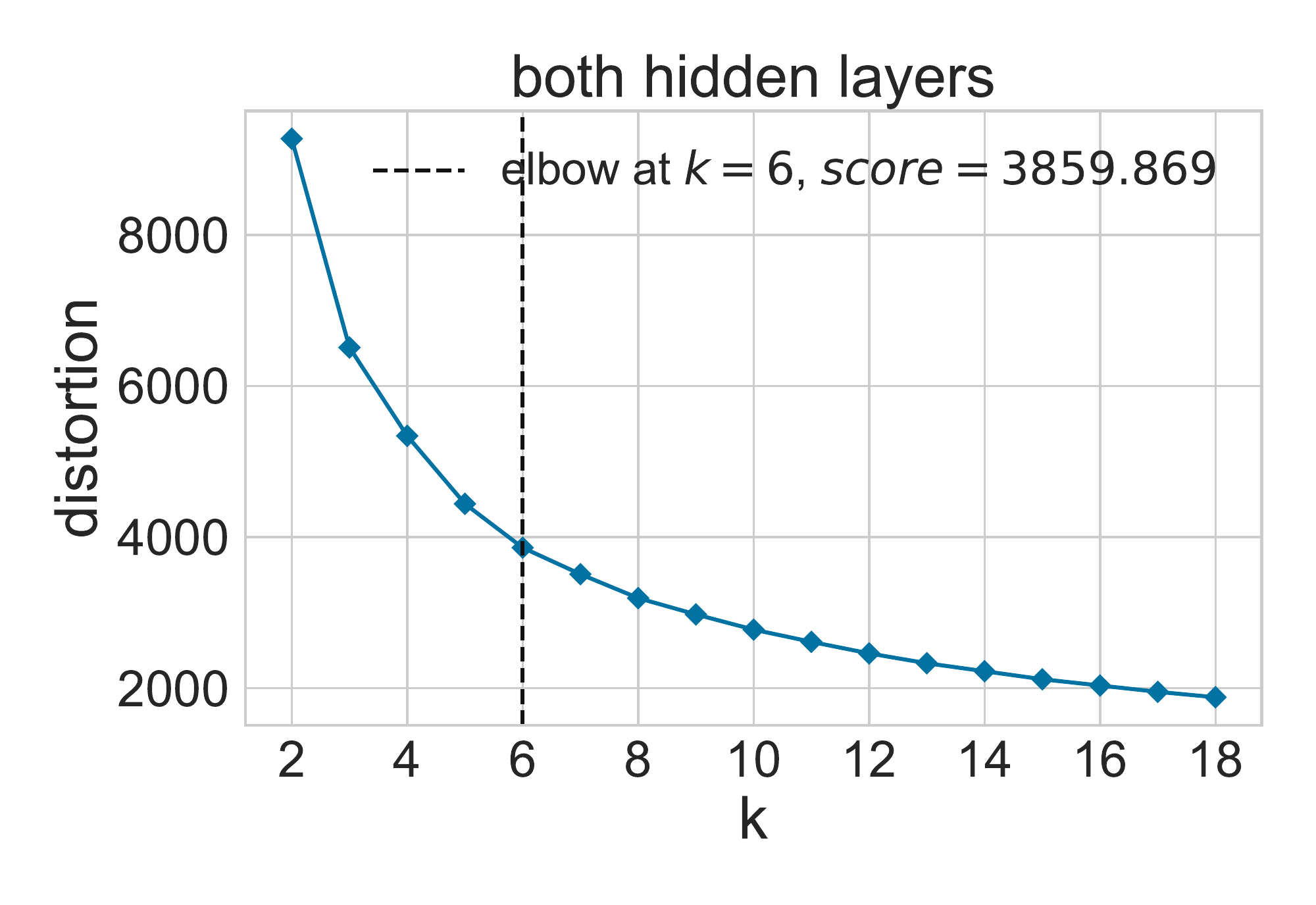}}  
     \subfigure[Fashion MNIST.]{\includegraphics[width=0.4 \linewidth]{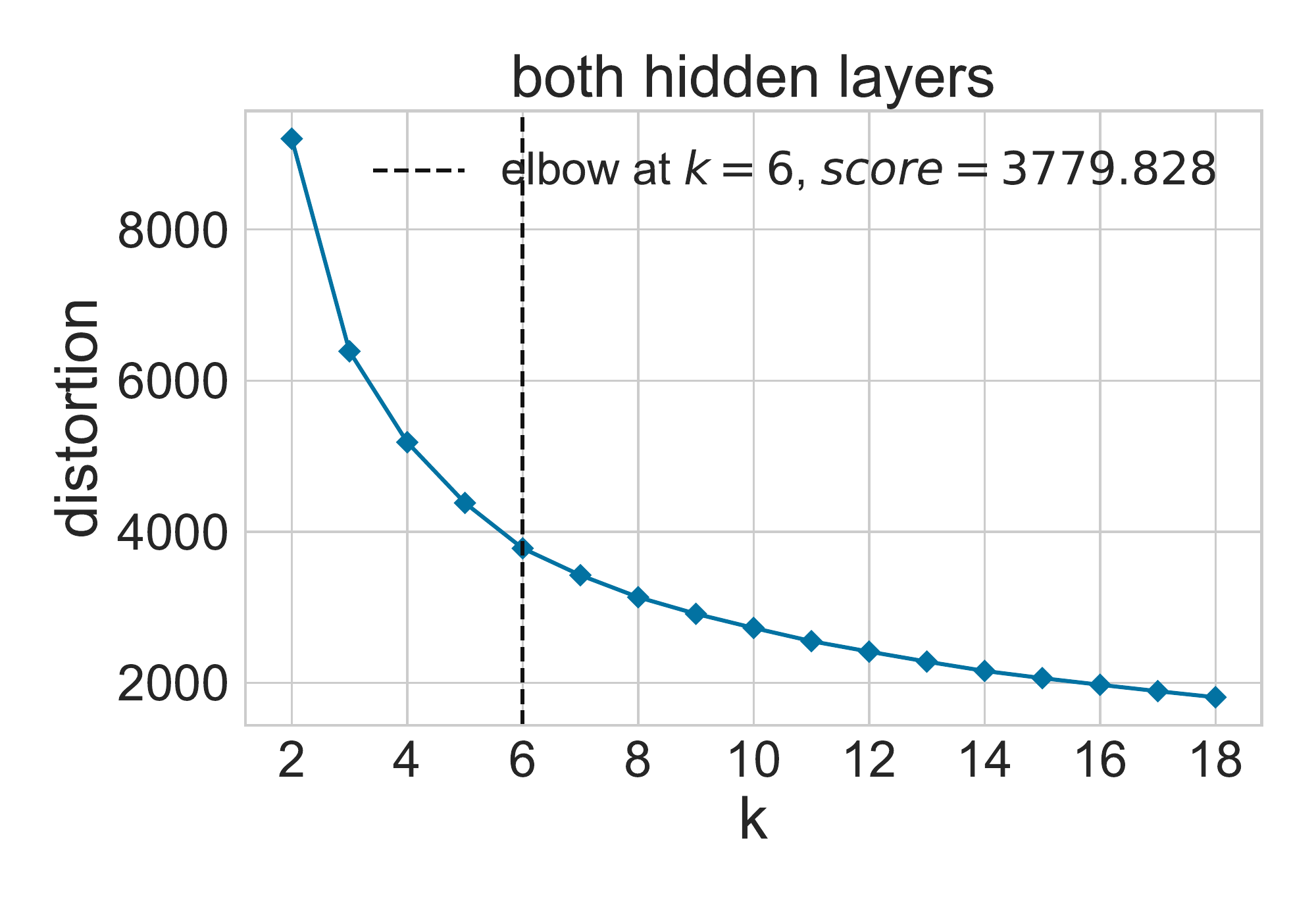}} \\
     \subfigure[CIFAR10.]{\includegraphics[width=0.4 \linewidth]{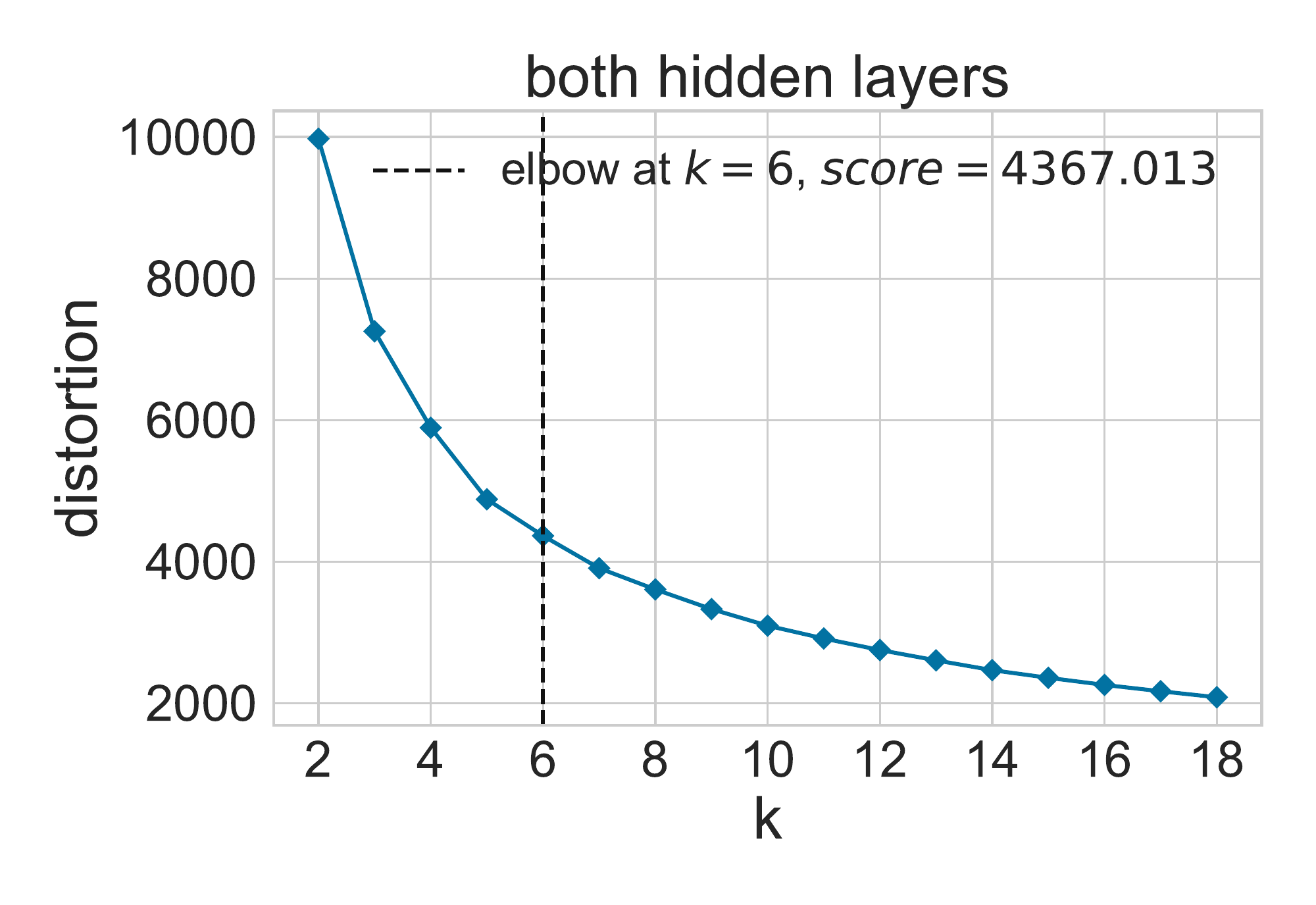}} \subfigure[KTH-TIPS.]{\includegraphics[width=0.4 \linewidth]{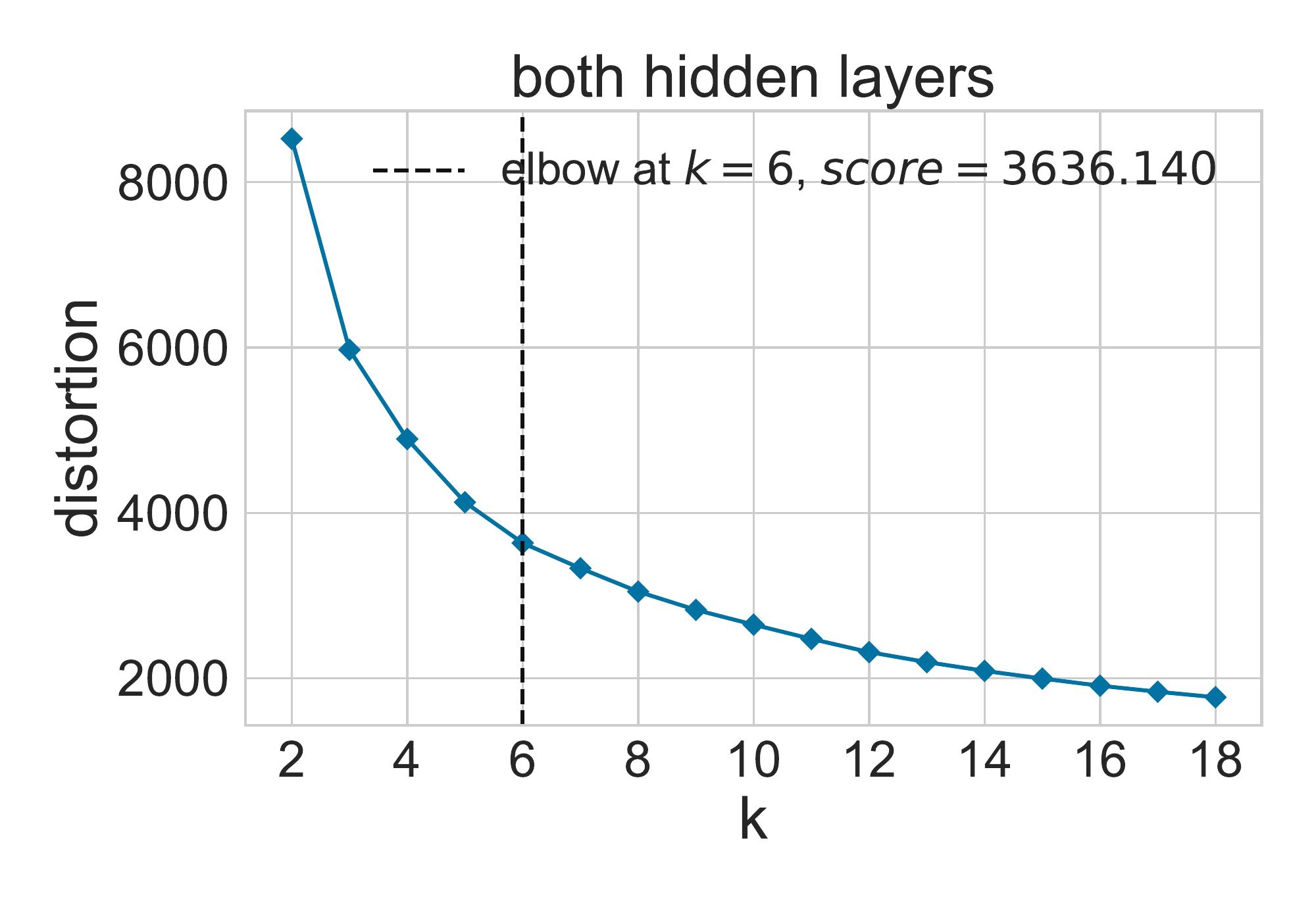}} 
    \caption{\label{fig:neurons_k}Finding the number of neuronal groups according to the Elbow method. Clustering is performed over hidden neuron's CN measures considering 1000 neural networks trained on each benchmark. Six neuron types are found independently of the considered domain or hidden layer.}
\end{figure}

The remaining 3 measures are considered for neural network characterization by building the local neuron descriptor $\psi(v_i) = [s(v_i), bc(v_i), sg(v_i)]$ and combine them to obtain matrix $D$ (see Equation \ref{eq:D}). Our BoN approach is then performed to obtain neuron groups in a non-supervised fashion using $k$-means. One of the drawbacks of $k$-means is that the parameter $k$ must be given beforehand, i.e., the number of data groups must be known, which is one of the main problems in data clustering regardless of the chosen algorithm. In our case, we cannot assume the number of possible neuron types in neural networks according to CN principles, nor did any previous work. Therefore, we use the elbow method for estimating the optimal number of groups which is a heuristic based on the distortion measure obtained from various $k$ values. This measure sums squared distances from each sample point to the nearest centroid found by $k$-means. Normally, increasing the number of clusters will naturally reduce the distortion since there are more clusters to use, but at some point, it is overfitting. The elbow method then estimates the point of inflection on the curve so that adding another cluster doesn't give much better modeling of the data. We tested $k$ values from 2 to 18, results are shown in Figure \ref{fig:neurons_k}. It is possible to observe that the elbow method indicates optimal $k$ around 6 in any case, either when grouping neurons from the first or second hidden layer separately or combined. It also happens independently of the dataset on which the neural networks were trained. This indicates an universal behavior of six topological signatures emerging on neurons within these networks, in different scenarios.

Considering the $k$-means results using $k=6$ over matrix $D$ with all hidden neurons, the obtained centroids $\psi_{i}$ on each dataset are shown on Table \ref{table:centroids_allnet}, sorted by strength ($s$). Another interesting pattern is observed here. The six neuron types detected are very similar across tasks, mainly when looking at strength and bipartite clustering. There are some particularities where measures vary from other cases, for instance, the strength and subgraph centrality on CIFAR10 at neuron type $\psi_{6}$, but most of the remaining properties are similar. Differences are expected in some cases, mainly on the first hidden layer, due to the natural differences in the input image patterns. For instance, MNIST represents digits covering a fraction of the image, while the remaining pixels are background. Fashion MNIST has a similar structure, with multiple background pixels. On the other hand, CIFAR10 and KTHTIPS has grayscale pixels covering the whole image (in most cases). The amount of background pixels then causes neurons of the first hidden layer to become more or less inhibitory. It directly influences the number of synapses connected to background information. Nevertheless, this seems to have a small impact on the six neuron types we found, while most of them have similar topological properties across the different tasks. This behavior also points to global topological patterns emerging from neural networks with different initial synapses and trained on different domains.

\begin{table}
	\caption{\label{table:centroids_allnet} The properties of the six neuronal groups found on neural networks trained on the corresponding benchmarks.}
		\label{table:student}
		\centering
	
		\begin{tabular}{|c|c|c|c|c|c|c|} 
        \hline
        MNIST &$\mathbf{\psi_{1}}$&$\mathbf{\psi_{2}}$&$\mathbf{\psi_{3}}$&$\mathbf{\psi_{4}}$&$\mathbf{\psi_{5}}$&$\mathbf{\psi_{6}}$\\
          \hline
        
        $\mathbf{s}$& -0.40 & -0.21 & -0.05 & 0.05 & 0.12 & 0.32\\
        $\mathbf{bc}$& 0.57 & 0.73 & 0.82 & 0.87 & 0.86 & 0.89\\
        $\mathbf{sg}$& 0.11 & 0.13 & 0.13 & 0.36 & 0.13 & 0.17\\
        
        \hline \hline
        
          Fashion MNIST &$\mathbf{\psi_{1}}$&$\mathbf{\psi_{2}}$&$\mathbf{\psi_{3}}$&$\mathbf{\psi_{4}}$&$\mathbf{\psi_{5}}$&$\mathbf{\psi_{6}}$\\
          \hline
      
        $\mathbf{s}$& -0.42 & -0.24 & -0.08 & 0.04 & 0.10 & 0.31\\
        $\mathbf{bc}$& 0.56 & 0.72 & 0.81 & 0.86 & 0.85 & 0.88\\
        $\mathbf{sg}$& 0.07 & 0.08 & 0.08 & 0.35 & 0.09 & 0.13\\

        \hline \hline
        
         CIFAR10 &$\mathbf{\psi_{1}}$&$\mathbf{\psi_{2}}$&$\mathbf{\psi_{3}}$&$\mathbf{\psi_{4}}$&$\mathbf{\psi_{5}}$&$\mathbf{\psi_{6}}$\\
          \hline

$\mathbf{s}$& -0.43 & -0.26 & -0.11 & -0.06 & 0.05 & 0.27\\
$\mathbf{bc}$& 0.57 & 0.73 & 0.82 & 0.85 & 0.85 & 0.89\\
$\mathbf{sg}$& 0.16 & 0.17 & 0.16 & 0.40 & 0.17 & 0.24\\

\hline \hline
      
      KTH-TIPS &$\mathbf{\psi_{1}}$&$\mathbf{\psi_{2}}$&$\mathbf{\psi_{3}}$&$\mathbf{\psi_{4}}$&$\mathbf{\psi_{5}}$&$\mathbf{\psi_{6}}$\\
          \hline
      
      $\mathbf{s}$& -0.42 & -0.25 & -0.11 & -0.04 & 0.04 & 0.26\\
    $\mathbf{bc}$& 0.56 & 0.73 & 0.82 & 0.87 & 0.86 & 0.89\\
    $\mathbf{sg}$& 0.13 & 0.14 & 0.13 & 0.33 & 0.14 & 0.19\\
      
        \hline
        \end{tabular} 
        
\end{table}

\begin{figure}
    \centering
    \subfigure[MNIST.]{
     \includegraphics[width=0.4\linewidth]{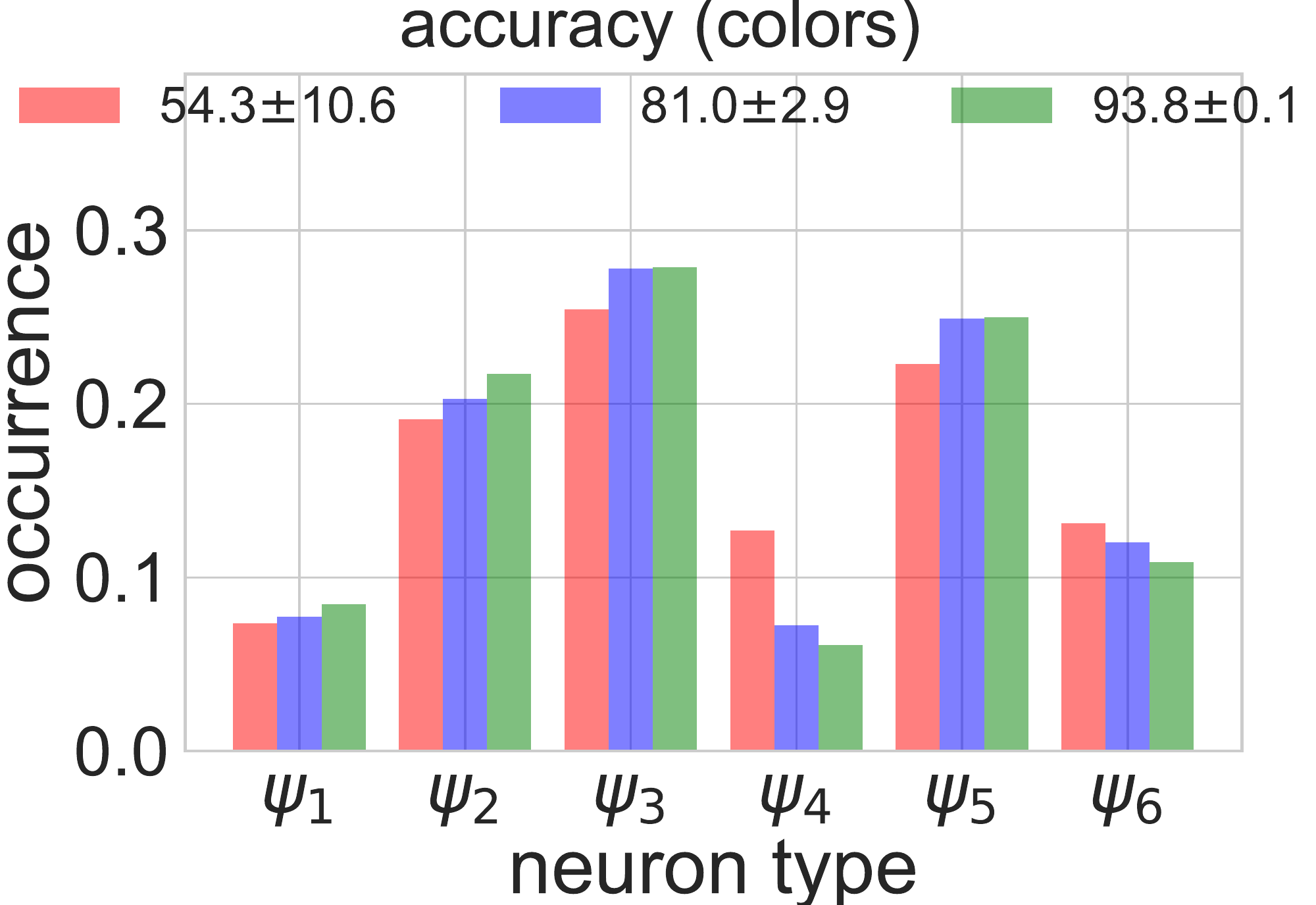}} \subfigure[Fashion MNIST.]{
     \includegraphics[width=0.4\linewidth]{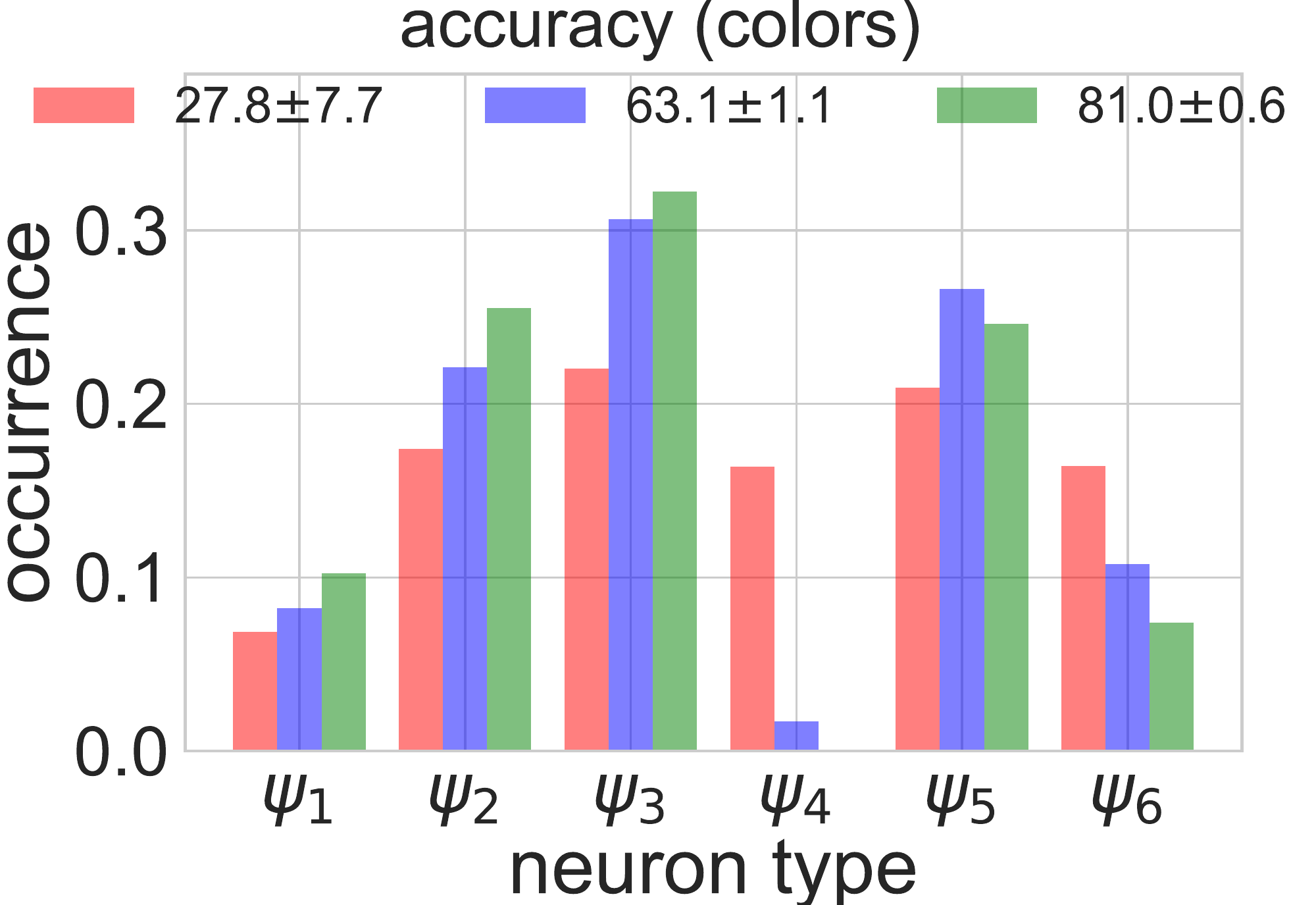}} 
     \subfigure[CIFAR10.]{
     \includegraphics[width=0.4\linewidth]{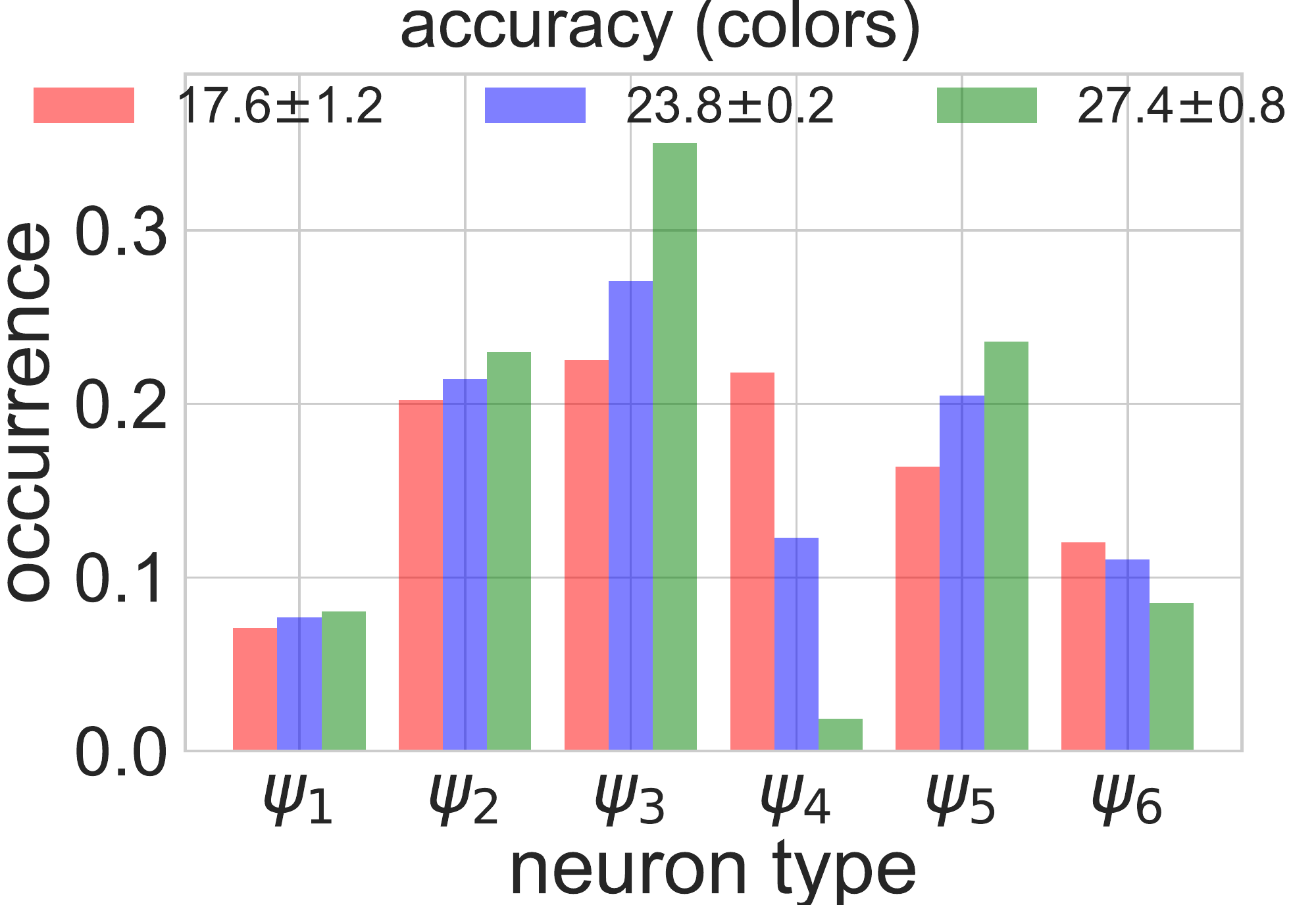}} \subfigure[KTH-TIPS.]{
     \includegraphics[width=0.4\linewidth]{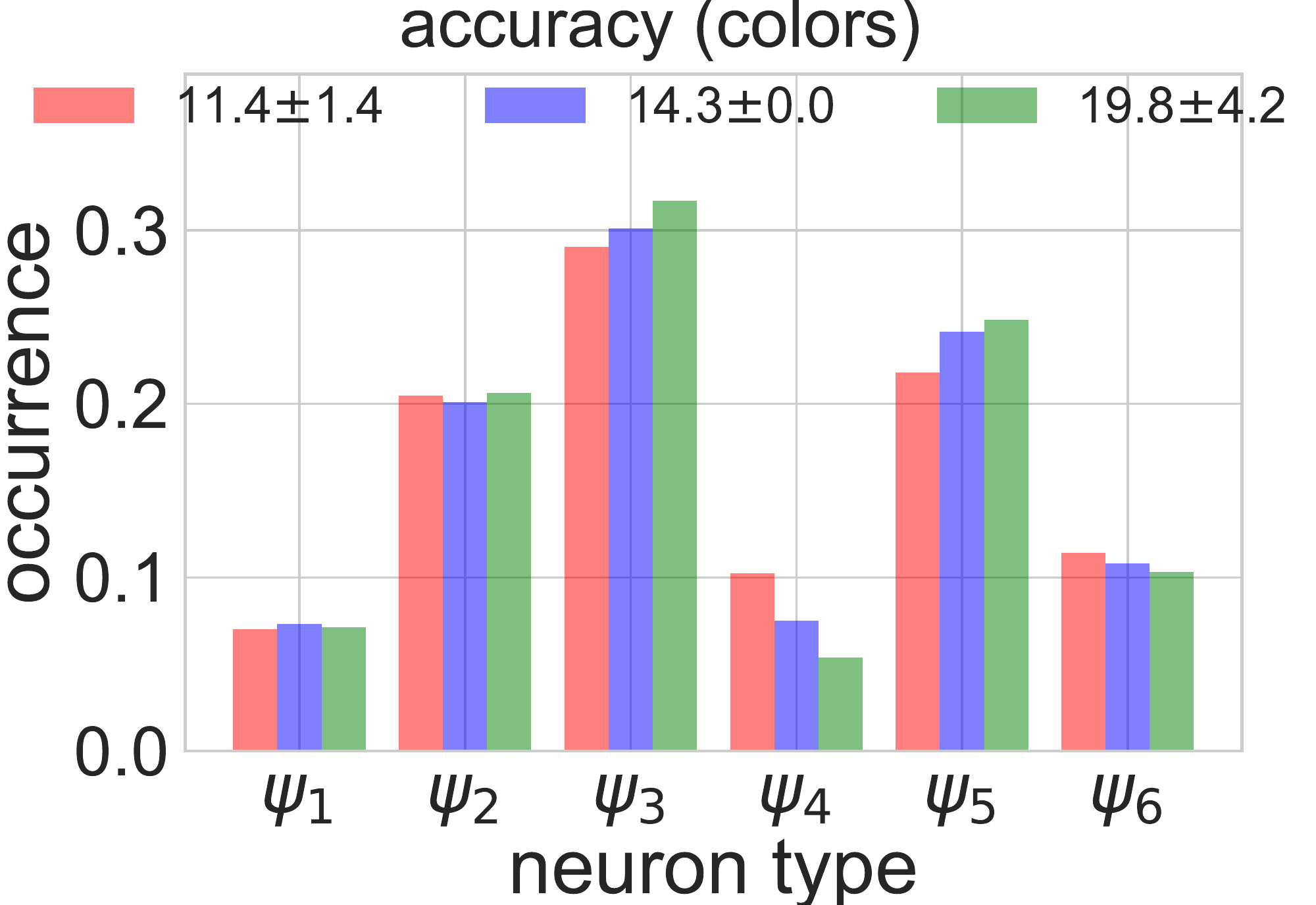}}
     
    \caption{\label{fig:neuronal_individual_dataset}Occurrence of neuron types ($\psi_i$) on neural networks trained on different vision benchmarks. Colors represent the 100 networks with the worst test accuracy (red), the 100 around the median (blue), and the top 100 (green). }  
\end{figure}

Using the obtained BoN is then possible to find neurons that match one of the six groups by simple nearest-neighbor association to its local descriptor. 
 To better compare the average structure of networks with different performance, we consider models from 3 accuracy ranges and compare their differences of neuron type occurrence. Figure \ref{fig:neuronal_individual_dataset} shows the distribution of neuron types for each benchmark the neural networks were trained. They are grouped according to the colors: red represents the 100 networks with lower accuracy, blue is the 100 around the median, and green the top 100. We can see some similarities across different tasks again. For instance, the occurrence of most neuron types follows a similar shape. These findings aligns with the results of \cite{testolin2020deep}, where authors have shown that receptive fields (neurons) with similar function develop similar topological properties in different target domains (they considered two benchmarks). It is widely known that receptive fields in deep neural networks tends to progressively extract higher-level features from the input. In vision, our study-domain, the first layers would then detect simple visual cues such as lines and circles, while deeper layers combine it into more detailed features. The simple features are usually universal, independently from the target domain.

The most interesting property we can observe within the neuron types occurrence is how the distributions change according to the neural network performance. The occurrence of most neuron types is clearly different for each accuracy range (colors). Another important finding is that neuron type $\psi_{3}$ is the most common among the best neural networks, having the highest occurrence and also highest occurrence difference to the lower performance networks. This group of neurons has a slight negative strength, closer to 0 compared to other neuron types. This phenomenon explains why weight initializers with normal distribution in small ranges around zero (such as Glorot \cite{glorot2010understanding}) tend to yield better results than widely uniform initializers. Combining with the other properties of this neuron type, we can assume that there is a strong tendency for neural networks with higher accuracy to contain around $30\%$ of hidden neurons with a slight negative strength $-0.1 \leq s \leq -0.05$, a considerably high clustering coefficient $bc \approx 0.8$, and subgraph-centrality $0.08 \leq sg \leq 0.16$. On the other hand, neuron type $\psi_{4}$ seems to be less critical for better-performing networks, and it appears more frequently among those with worse performance. These neurons have a small strength magnitude and one of the highest clustering coefficients. However, their most interesting property is the highest subgraph centrality than all other types, by a significant margin, and this is also observed throughout different benchmarks. It is a notable contrast and highlights the difference between structure and performance on these networks. We then conclude that neural networks with poor performance tend to have considerably more neurons with higher subgraph centrality ($sg > 0.3$) when compared to other networks. This property may be highlighting neuron saturation, usually caused by fading/exploding gradients during gradient descent, which would explain why these networks fail to achieve higher performance.

\begin{table*}[!htb]
 \caption{The Jensen-Shannon Divergence between pairs of neuron type frequency from BoN vocabularies obtained in the same or different benchmarks, averaged over all 1000 samples in each case (standard deviation in brackets).}\label{tab:crossoccurrence}
    \centering
    \begin{tabular}{|cc|c|c|c|c|}
    \hline
            &  & \multicolumn{4}{c|}{\textbf{target}} \\
            
            &&MNIST & Fashion MNIST & CIFAR10 & KTH-TIPS \\
           \multirow{4}{*}{\rotatebox[origin=c]{90}{\parbox[c]{1.2cm}{\centering \textbf{vocabulary}}}}& MNIST & 0 & 0.21 ($\pm$ 0.16)& 0.32 ($\pm$ 0.17) & 0.27 ($\pm$ 0.14)\\
            &Fashion MNIST& 0.24 ($\pm$ 0.17) & 0 & 0.33 ($\pm$ 0.2)& 0.28 ($\pm$ 0.17) \\
            &CIFAR10&0.29 ($\pm$ 0.16) & 0.25 ($\pm$ 0.18)&0&0.24 ($\pm$ 0.19)\\
            &KTH-TIPS& 0.27 ($\pm$ 0.15) & 0.24 ($\pm$ 0.16) & 0.28 ($\pm$ 0.19) &0\\
       
          \hline
          
    \end{tabular}

\end{table*}

We also calculate the similarities in neuron type occurrence when obtaining the BoN in different benchmarks. The Jensen-Shannon Divergence (JSD) is considered, which provides a smoothed and normalized version of the Kullback-Leibler Divergence \cite{kullback1951information} (KLD). This metric quantifies the difference (or similarity) between two probability distributions, with scores between 0 (identical) and 1 (maximally different). We compute neuron types occurrence on networks for BoNs obtained in the same benchmark the model was trained ($\psi$), or in a different benchmark ($\bar{\psi}$). We then calculate $KLD(\psi,\bar{\psi}) = \sum_i p(\psi_i) \log(\frac{p(\psi_i)}{p(\bar{\psi_i})})$, which is used for computing $JSD = \frac{1}{2} (KLD(\psi, \frac{\psi + \bar{\psi}}{2}) + KLD(\bar{\psi}, \frac{\psi + \bar{\psi}}{2}))$. The JSD is then averaged over frequency pairs of all 1000 networks, for all possible combinations between benchmarks for construction and application of the BoN, results are shown on Table \ref{tab:crossoccurrence}. Despite the high standard deviation, the average divergence values are relatively low, around $0.21$ and $0.33$, which indicates significant similarity between neuron types. Nevertheless, these results suggest an average behavior of similarity emerging within neuron types from BoNs obtained with neural networks trained on different benchmarks.

\section{Conclusion}\label{sec:conclusion}

In this work, we explore the correlations between the structure and performance of fully connected neural networks on large vision tasks. Our first contribution is a new dataset with 4 thousand neural networks, each with different initial random synapses, applied on known vision benchmarks (MNIST, Fashion MNIST, CIFAR10, and KTH-TIPS). We then consider each neural network as weighted and undirected graphs and focus on analyzing local neuron properties. Eight CN centrality measures are computed for each hidden neuron, which is also included in the dataset. Our analysis shows that some of them (average nearest neighbor strength, current-flow closeness, and harmonic centrality) do not present relevant information regarding the model's performance. We also find redundancy in some cases, where correlations were high (second-order centrality and number of cliques). In the end, three centrality measures contain the most relevant information (strength, bipartite clustering, and subgraph centrality) and show interesting neural network properties. 

Our results show a strong correlation between neuronal centrality properties and neural network performance. In some cases, it is possible to linearly separate the top and worst 100 neural networks by simply taking the hidden layers average of a single CN measure. We also propose a local descriptor for hidden neurons by concatenating the three best measures and found that they are usually grouped in 6 neuron types throughout the neural network population. Each group has a distinct topological signature and share similarities even between the different tasks the neural networks are employed. Some neuron types are more common than others, and their distribution seem to be directly related to performance. 
We were able to highlight the CN properties related to this phenomenon, with distinct differences between different performing networks. For instance, neurons wit a combination of high clustering coefficient, and, most importantly, a distinct higher subgraph centrality tends to emerge more frequently on lower performing models.

Network science is a strong candidate to understand the internal properties of neural networks. Our results shows that the structure and performance of fully connected models seem to be directly related. Our findings are important for understanding the functioning of neurons, their learning dynamics, and their impacts on the whole system performance. We will consider these CN properties dynamically for future works, i.e., how they emerge from random initial weights during training and the impacts of different initial structures. Moreover, to analyze the system stability, such as the impacts of adversarial samples. And finally, to target practical applications of these principles, such as employing CN properties for better neural network construction, stability, and learning.

\section*{Acknowledgments}

Leonardo F. S. Scabini acknowledges funding from the São Paulo Research Foundation (FAPESP) (Grant \#2019 /07811-0) and CNPq (Grant \#142438/2018-9). Odemir M. Bruno acknowledges funding from CNPq (Grant \#307897/2018-4) and FAPESP (grant \#2014/08026-1 and 2016/18809-9). The authors are also grateful to the NVIDIA GPU Grant Program for donating a Titan Xp GPU used in this research.

\bibliographystyle{unsrt}
\bibliography{arxiv.bib}

\end{document}